\def\year{2023}\relax
\documentclass[letterpaper]{article} 
\usepackage{aaai23}  
\usepackage{times}  
\usepackage{helvet}  
\usepackage{courier}  
\usepackage[hyphens]{url}  
\usepackage{graphicx} 
\usepackage{natbib}  
\usepackage{caption} 
\usepackage{algorithm}
\usepackage{algorithmic}
\usepackage{subfigure}
\input{lc_commands.tex}
\usepackage{multirow}
\usepackage{pifont}
\newcommand{\model}{PMMM}

\usepackage{newfloat}
\usepackage{listings}
\DeclareCaptionStyle{ruled}{labelfont=normalfont,labelsep=colon,strut=off} 
\floatstyle{ruled}
\newfloat{listing}{tb}{lst}{}
\floatname{listing}{Listing}
\nocopyright

\title{Differentiable Meta Multigraph Search with Partial Message Propagation \\on Heterogeneous Information Networks}
\author{
    Chao Li\textsuperscript{\rm 1}$^,$\textsuperscript{\rm 2}$^,$\equalcontrib,
    Hao Xu\textsuperscript{\rm 1}$^,$\textsuperscript{\rm 2}$^,$\equalcontrib,
    Kun He\textsuperscript{\rm 1}$^,$\textsuperscript{\rm 2}$^,$\thanks{Corresponding author.}
}
\affiliations{
    \textsuperscript{\rm 1}School of Computer Science, Huazhong University of Science and Technology, Wuhan 430074, China.\\ 
    \textsuperscript{\rm 2}Hopcroft Center on Computing Science, Huazhong University of Science and Technology, Wuhan 430074, China.\\
    \{D201880880, M202173589, brooklet60\}@hust.edu.cn 
}

\usepackage{bibentry}

\begin{document}

\maketitle

\begin{abstract}
Heterogeneous information networks (HINs) are widely employed for describing real-world data with intricate entities and relationships. To automatically utilize their semantic information, graph neural architecture search has recently been developed on various tasks of HINs. Existing works, on the other hand, show weaknesses in instability and inflexibility.
To address these issues, we propose a novel method called \textit{Partial Message Meta Multigraph search} (PMMM) to automatically optimize the neural architecture design on HINs. Specifically, to learn how graph neural networks (GNNs) propagate messages along various types of edges, PMMM adopts an efficient differentiable framework to search for a meaningful meta multigraph, which can capture more flexible and complex semantic relations than a meta graph. The differentiable search typically suffers from performance instability, so we further propose a stable algorithm called partial message search to ensure that the searched meta multigraph consistently surpasses the manually designed meta-structures, \ie, meta-paths. Extensive experiments on six benchmark datasets over two representative tasks, including node classification and recommendation, demonstrate the effectiveness of the proposed method. Our approach outperforms the state-of-the-art heterogeneous GNNs, finds out meaningful meta multigraphs, and is significantly more stable. 
\end{abstract}

\section{Introduction}
\label{sec:intro}
Heterogeneous information networks (HINs) are wide-spread in the real world for abstracting and modeling complex systems 
for the
rich semantic information
provided by heterogeneous relations consisting of multiple node types, edge types, and node features.  
For instance, bibliographic graphs (such as DBLP) with three types of nodes, \ie, author, paper, and venue, include multiple edge types, such as co-authorships, co-citations, publishing in the same venue.

To model the heterogeneous structure, there have been numerous 
heterogeneous GNNs combining meta-paths to learn with HINs~\cite{DBLP:conf/esws/SchlichtkrullKB18,DBLP:conf/kdd/ZhangSHSC19,DBLP:conf/www/WangJSWYCY19,gt}.
More recently, in light of the achievement of neural architecture search (NAS) in convolutional neural networks (CNNs), several works have extended NAS to heterogeneous GNNs, based on the ability to automate the customization of neural architecture for specific datasets and tasks.

However, there are some weaknesses in existing NAS methods for heterogeneous GNNs. 
For instance, GEMS~\cite{han2020genetic} uses an  evolutionary algorithm as the search strategy, making its search cost dozens of times 
as 
training a single GNN. 
HGNAS~\cite{gao2021heterogeneous}, employing reinforcement learning as the search strategy, has the same issue of inefficiency. 
Inspired by the simplicity and computational efficiency of differentiable architecture search~\cite{liu2018darts} in CNNs, DiffMG~\cite{DBLP:conf/kdd/DingYZZ21} proposes to search a meta graph in a differentiable fashion, making the search cost on a par with training a single GNN once. Yet, it suffers from performance instability and often finds architectures worse than hand-designed models. 

Besides, existing works have a common weakness of inflexibility. For example, GEMS~\cite{han2020genetic} and DiffMG~\cite{DBLP:conf/kdd/DingYZZ21} search for meta graphs to guide how GNNs propagate messages. However, meta-paths and meta graphs are initially defined for hand-designed heterogeneous GNNs with fixed patterns, so they are insufficient to encode various rich semantic information on diverse HINs and will restrict the searched architecture to inflexible topology. HGNAS~\cite{gao2021heterogeneous} searches for message encoding and aggregation functions but propagates messages using predefined meta-paths or meta graphs.  
It motivates us to propose a more expressive meta-structure for NAS on HINs.  


\begin{figure*}[tb]
    \centering
    \subfigure[An example academic network]{\includegraphics[width=0.32\linewidth]{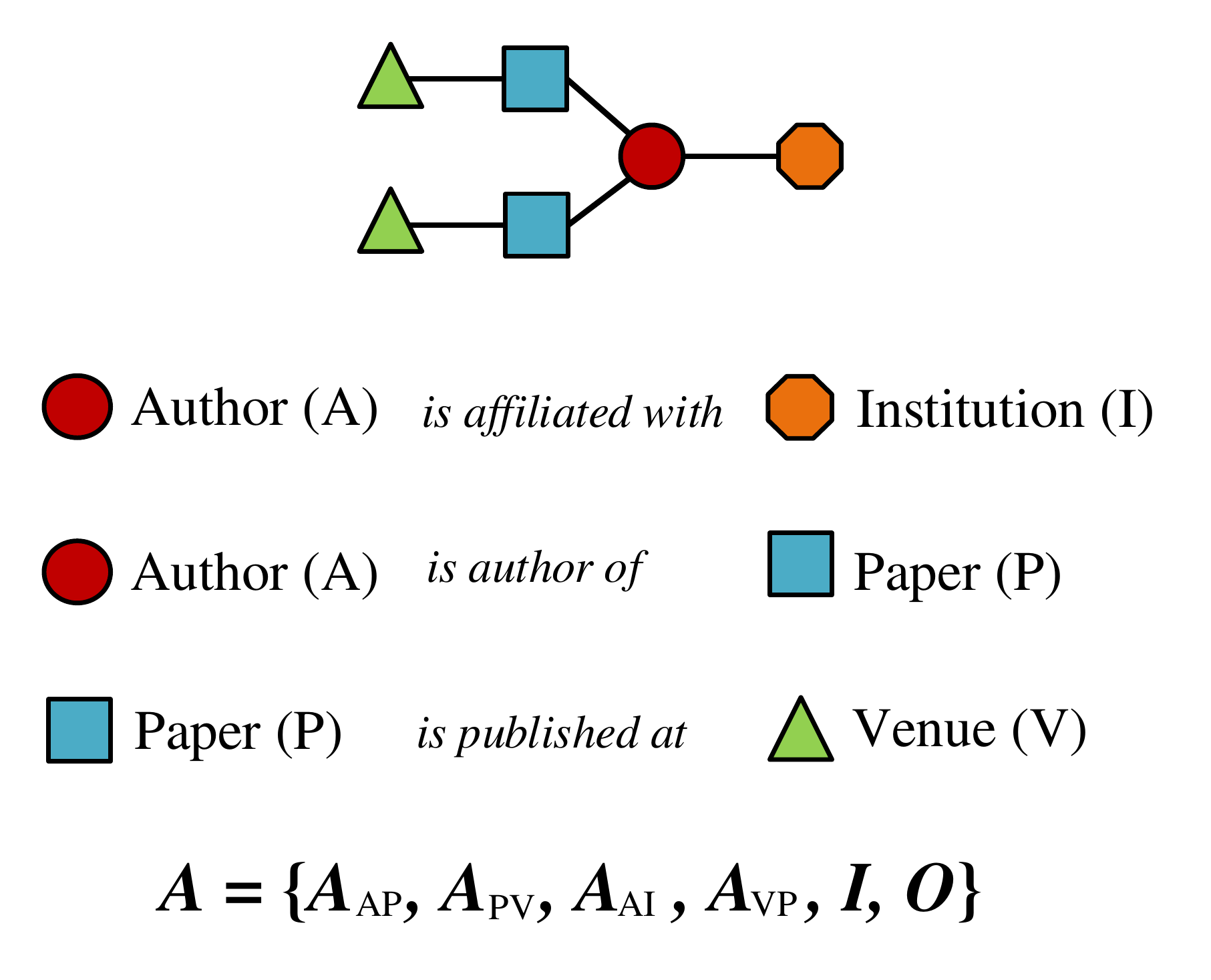}}
    \subfigure[Meta graph]{\includegraphics[width=0.33\linewidth]{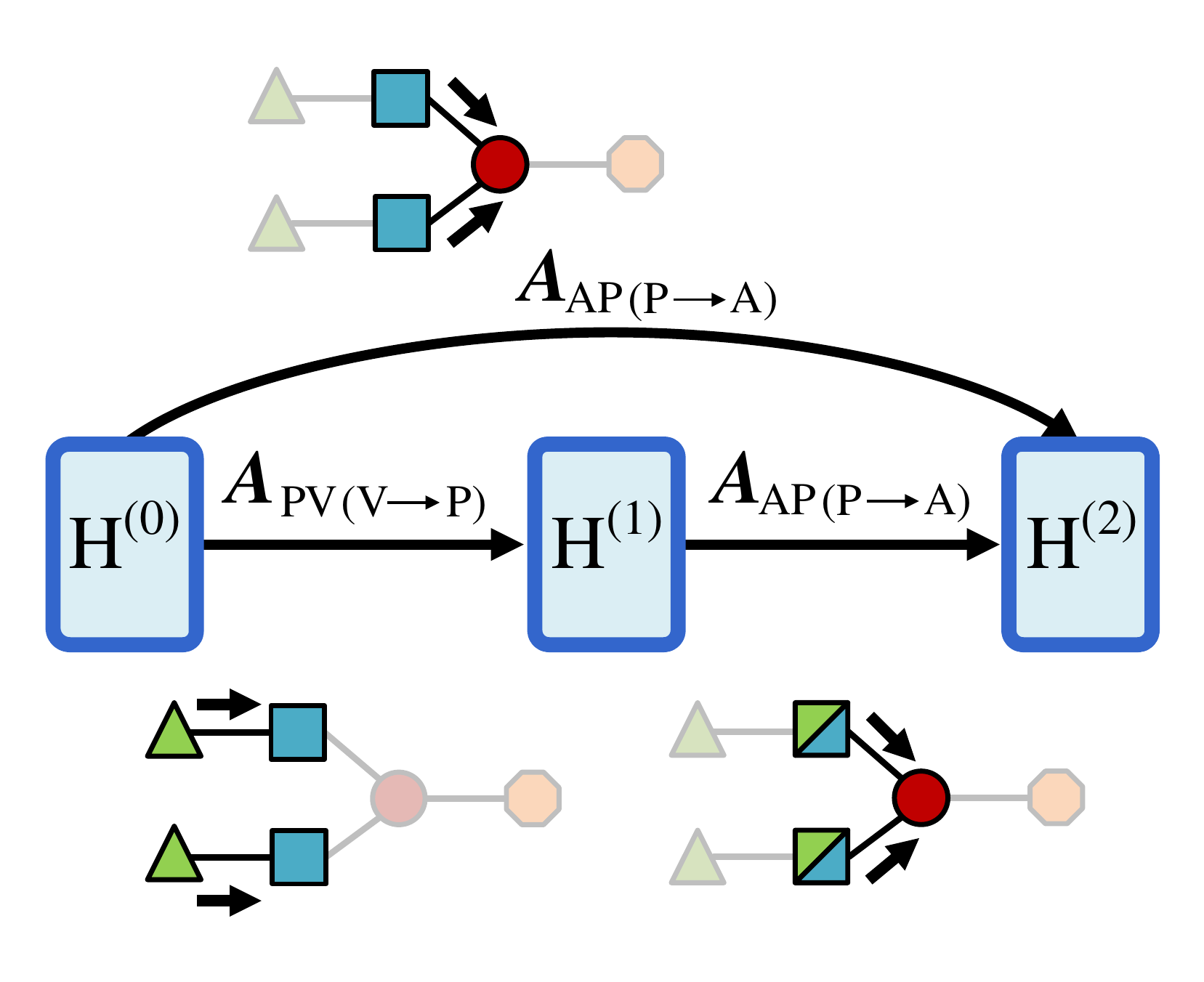}}
    \subfigure[Meta  multigraph]{\includegraphics[width=0.33\linewidth]{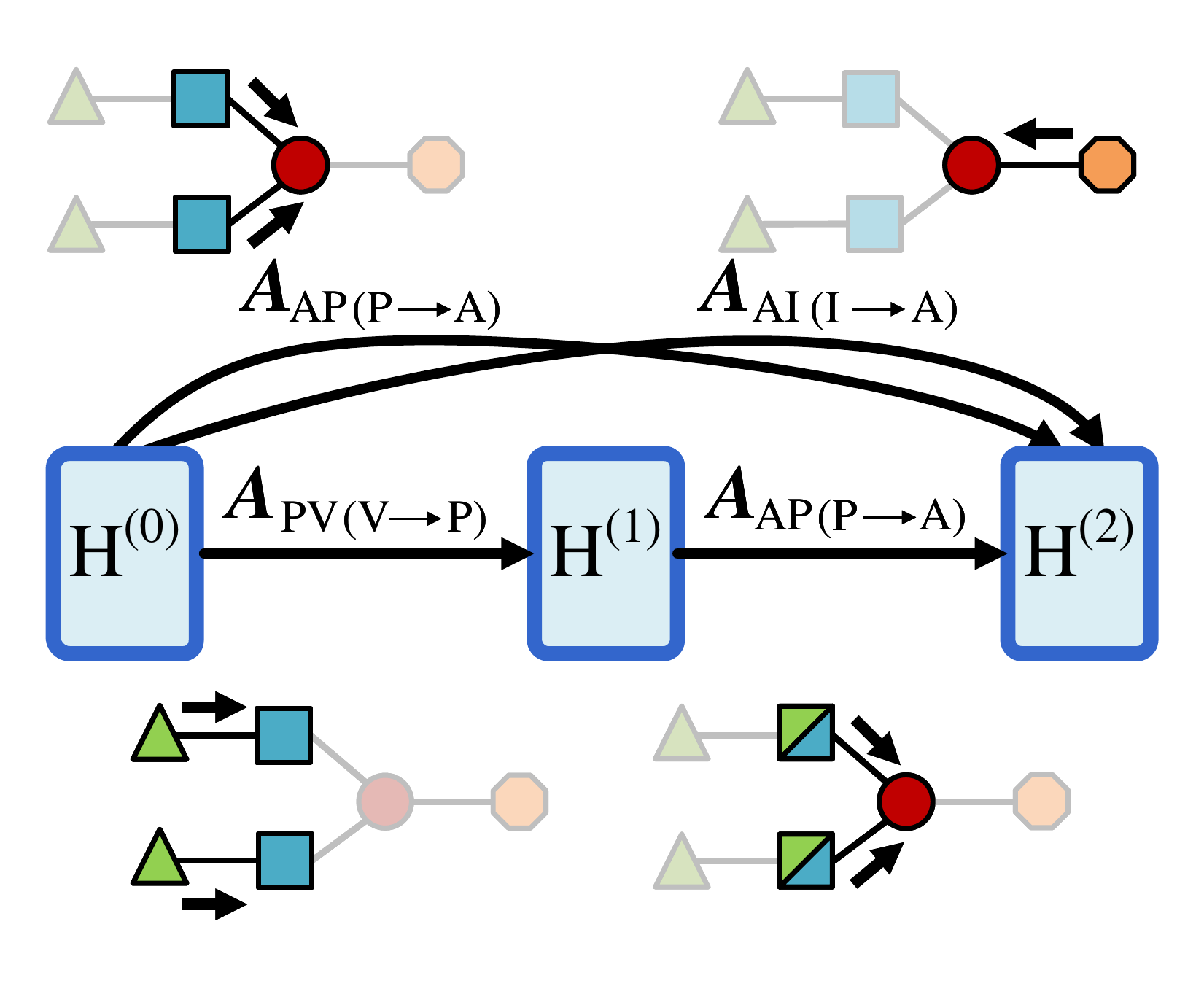}}
    \vspace{-2pt}
    \caption{Illustration of (a) an example academic network, (b) meta graph, and (c) meta multigraph with depth $N=2$. $\bm{A}$ is the set of candidate message passing types. $\bm{I}$ and $\bm{O}$ stand for the identity matrix and empty matrix,  respectively. A meta multigraph allows propagating multiple types of message between nodes, which can not be realized by a meta graph.
    }
    \vspace{-10pt}
    \label{fig:meta-structure}
\end{figure*}
To this end, we present a new method called PMMM (Partial Message Meta Multigraph search), which is composed of a novel differentiable search  algorithm for stable search, namely partial message search, as well as an innovative meta-structure for flexible topology, namely meta multigraph. 
Specifically, to stabilize the differentiable search, PMMM randomly selects partial candidate message passing steps to update per iteration to ensure that all candidate paths are equally and fully trained, as well as to decouple the joint optimization of paths.
To derive flexible topology, PMMM searches for a novel meta multigraph by selecting the top-$k$ most promising candidate message passing types for aggregation. Searching for a meta multigraph is a free performance-enhancing strategy as it can encode more flexible and sophisticated semantic information in HINs compared with a meta-path or meta graph.  
Experiments show that PMMM consistently outperforms state-of-the-art baselines and  significantly improves the stability compared with differential meta graph search on both node classification and recommendation tasks. 
Our main contributions are threefold: 
\begin{itemize}
\item To our knowledge, PMMM is the first NAS method to search multigraph as the architecture. 
We propose a new concept of meta multigraph 
to guide GNNs to propagate messages for NAS on HINs. 
\item We propose the first stable 
differentiable architecture search algorithm on HINs, called partial message search, 
which can consistently discover promising architectures that outperform hand-designed models.
\item Thorough experiments are conducted to demonstrate the effectiveness, stability, and flexibility of our method. 
\end{itemize}


\section{Related Works}
\label{sec:related_work}
\subsection{Heterogeneous GNNs}
Heterogeneous GNNs are proposed to handle HINs as 
rich and diverse information could be better utilized.
One category of heterogeneous GNNs uses hand-designed meta-paths to define neighbors, including HAN~\cite{wang2019heterogeneous}, MAGNN~\cite{fu2020magnn}, NIRec~\cite{jin2020efficient}, \etal 
Different from these methods, our PMMM does not require specific prior knowledge.
Another category aims to solve the meta-path selection conundrum via fusing various edge types based on the attention mechanisms, including GTN~\cite{NIPS2019_9367}, HetGNN~\cite{zhang2019heterogeneous}, HGT~\cite{hu2020heterogeneous}, \etal 
Compared with these methods, PMMM can find efficient architectures by filtering out unrelated edge types. 

\subsection{Graph Neural Architecture Search}
Neural architecture search (NAS) has shown promising results in searching for convolutional neural networks since it has brought up the prospect of automating the customization of neural architectures for specific tasks~\cite{zoph2016neural,xie2017genetic,liu2018darts}. 
Recently, numerous NAS-based works have been proposed to obtain data-specific homogeneous GNN topology~\cite{zhou2019auto,qin2021graph,zhao2021search,wei2022designing}. 
Meanwhile, only a few works 
attempt to employ NAS in HINs due to the complex semantic relationships. 
GEMS~\cite{han2020genetic} is the first NAS method on HINs that uses an evolutionary algorithm to search for meta graphs for recommendation.
HGNAS~\cite{gao2021heterogeneous} searches for message encoding and aggregation functions by using reinforcement learning. 
Considering the inefficiency of evolutionary algorithm and reinforcement learning, 
DiffMG~\cite{DBLP:conf/kdd/DingYZZ21} employs an differentiable algorithm to search for meta graphs. However, its performance is unstable. 
Compared with them, PMMM can perform an efficient and stable search for different tasks on HINs. Furthermore, our model searches for a meta multigraph, which shows better diversity and a stronger capacity to capture complex semantic information.

\section{Definitions}
We first provide several definitions used in the literature. 

\noindent\textbf{Definition 2.1} Heterogeneous Information Network (HIN)~\cite{sun2011pathsim}.
An HIN 
is defined as a directed graph $\mathcal{G}=\{\mathcal{V}, \mathcal{E}, \mathcal{T}, \mathcal{R}, f_{\mathcal{T}}, f_{\mathcal{R}}\}$, where $\mathcal{V}$ denotes the set of nodes and $\mathcal{E}$ denotes the set of edges, $\mathcal{T}$ is the node-type set and $\mathcal{R}$ is the edge-type set. Each node $v \in \mathcal{V}$ and each edge $e \in \mathcal{E}$ are associated with their type mapping functions $f_{\mathcal{T}}(v) \in \mathcal{T}$ and $f_{\mathcal{R}}(e) \in \mathcal{R}$, respectively. 
We define its network schema as 
$\mathcal{S}=\{\mathcal{T}, \mathcal{R}\}$, with $\lvert \mathcal{T} \rvert >1$ or $\lvert \mathcal{R} \rvert >1$.


\noindent\textbf{Definition 2.2} Meta-path~\cite{Metapath}. A meta-path $P$ is a path with length $l$ defined on the schema $\mathcal{S}=\{\mathcal{T}, \mathcal{R}\}$ of $\mathcal{G}=\{\mathcal{V}, \mathcal{E}, \mathcal{T}, \mathcal{R}, f_{\mathcal{T}}, f_{\mathcal{R}}\}$, and is denoted in the form of $t_1\stackrel{r_1}{\longrightarrow}t_2\stackrel{r_2}{\longrightarrow}...\stackrel{r_l}{\longrightarrow}t_{l+1}$, where $t_1,...,t_{l+1} \in \mathcal{T}$ and $r_1,...,r_l \in \mathcal{R}$. 
One meta-path can correspond to multiple meta-path instances in the underlying HIN.


\noindent\textbf{Definition 2.3} Meta Graph. A meta graph
is a directed acyclic graph on the network schema $\mathcal{S}$ with a single source node $t_s \in \mathcal{T}$ (with zero in-degree) and a single sink (target) node 
$t_e \in \mathcal{T}$ (with zero out-degree). 

As shown in Fig.~\ref{fig:meta-structure} (b), a meta graph can only propagate one message passing type between two nodes, which is insufficient to encode rich semantic information on HINs. One extreme example is when all candidate message passing types are necessary, a meta graph will have extremely bad performance due to the restriction of only retaining one message passing type.
Another example is shown in Fig.~\ref{fig:meta-structure} (c). The meta multigraph in (c) allows the author to aggregate the initial information of both paper and institution, while the meta graph in (b) can only aggregate the initial information of either paper or institution. Based on the above concerns, we define the meta multigraph to facilitate the  description of our method. 

\noindent\textbf{Definition 2.4} Meta Multigraph. A meta multigraph defined on network schema $\mathcal{S}$ is a directed acyclic multigraph consisting of multi-edges and hyper-nodes. Each multi-edge $\mathcal{R}_i \subseteq \mathcal{R}$ consists of multiple edges. Each hyper-node $\mathcal{T}_j \subseteq \mathcal{T}$ is the set of the heads of its incoming multi-edges and the tails of its outgoing multi-edges. A meta multigraph contains a single source hyper-node and a single sink (target) hyper-node. 



A meta multigraph allows propagating multiple message passing types between two different hyper-nodes, offering a natural generalization of a meta graph. 



\section{Methodology}
\label{sec:method}
In this section, we first briefly show the framework of the proposed differentiable meta multigraph search on HINs, then we develop a partial message search algorithm and meta multigraph deriving strategy, showing how they improve the stability and generate flexible and effective meta multigraphs, respectively.

\subsection{Differentiable Meta Multigraph Search}
\label{m1}

Differentiable meta multigraph search is developed on differentiable meta graph search  (DiffMG)~\cite{DBLP:conf/kdd/DingYZZ21}. The training of differentiable meta multigraph search consists of two stages. At the search stage, we train a super-net, from which sub-networks can be sampled exponentially. The super-net is constructed by 
directed acyclic multigraphs, whose multi-edges consist of multiple paths corresponding to the candidate message passing types. 
Each candidate message passing type is weighted by architecture parameters, which are jointly optimized with the super-net weights in an end-to-end manner. The goal of the search stage is to determine the architecture parameters. 
At the evaluation stage, the strongest sub-network is preserved as the target-net by pruning redundant paths based on the searched architecture parameters. The target-net is then retrained from scratch to get the final results. 

We define the search space of differentiable meta multigraph search as a directed acyclic multigraph, in which the ordered nodes $\bm{H}=\{\bm{H}^{(0)},\bm{H}^{(1)},\cdots,\bm{H}^{(n)},\cdots,\bm{H}^{(N)}\}$ denote the hyper-nodes in the message passing process and the multi-edges $E= \{\mathcal{R}^{(i,j)}|0 \le i<j\le N\}$ signify the message passing types between hyper-nodes. Each hyper-node $\bm{H}^{(n)} \subseteq \mathcal{T}$. $\bm{H}^{(0)}$ and $\bm{H}^{(N)}$ is the input and output of the meta multigraph, respectively. 
Each multi-edge $\mathcal{R}^{(i,j)} \subseteq \mathcal{R}$ contains multiple paths, corresponding to candidate message passing types.  
For $r \in \mathcal{R}^{(i,j)}$, we use  $\bm{\mathcal{A}}^{(i,j)}_{r}$ to denote the adjacency matrix formed by the edges of type $r$ in $\mathcal{G}$. 
The core idea of differentiable meta multigraph search is to formulate the information propagated from $\bm{H}^{(i)}$ to $\bm{H}^{(j)}$ as a weighted sum over all candidate message passing steps, namely: 

\begin{align}
\bm{H}^{(j)}=\sum_{i<j}\sum_{r \in \mathcal{R}^{(i,j)}} p^{(i,j)}_r {f}\left(\bm{\mathcal{A}}^{(i,j)}_r,\bm{H}^{(i)}\right),
\label{eq:forward}
\end{align}
\begin{align}
p^{(i,j)}_r=\exp(\alpha^{(i,j)}_r)/\sum_{r \in \mathcal{R}^{(i,j)}}\exp(\alpha^{(i,j)}_r).
\label{eq:softmax}
\end{align}
Here ${f}(\bm{\mathcal{A}}^{(i,j)}_r,\bm{H}^{(i)})$ denotes one message passing step that aggregates
$\bm{H}^{(i)}$ along   $\bm{\mathcal{A}}^{(i,j)}_r$, $\alpha^{(i,j)}_r$ indicates the architecture parameters of 
$\bm{\mathcal{A}}^{(i,j)}_r$, and $p^{(i,j)}_r \in (0,1]$ denotes the corresponding path strength calculated by a softmax over $\alpha^{(i,j)}_r$. ${f}(\cdot)$ can be any aggregation function. Following DiffMG~\cite{DBLP:conf/kdd/DingYZZ21}, we employs the aggregation function in graph convolutional network (GCN).

The parameter update in Eq.~\ref{eq:forward} involves a bilevel optimization problem~\cite{anandalingam1992hierarchical, colson2007overview,xue2021rethinking}: 
\begin{align}
	\min_{\bm{\alpha}} \quad & \mathcal{L}_{val}(\bm{\omega}^*(\bm{\alpha}), \bm{\alpha}) \label{eq:outer} \\
	\text{s.t.} \quad &\bm{\omega}^*(\bm{\alpha}) = \mathrm{argmin}_{\bm{\omega}} \enskip \mathcal{L}_{train}(\bm{\omega}, \bm{\alpha}) ,
	\label{eq:inner}
\end{align}
where $\mathcal{L}_{train}$ and $\mathcal{L}_{val}$ denote the training and validation loss, respectively. The goal of the search stage is to find $\bm{\alpha}^*$ that minimizes $\mathcal{L}_{val}(\bm{\omega}^*, \bm{\alpha}^*)$. 

At the evaluation stage, we derive a compact meta multigraph by pruning redundant paths based on path strengths $p^{(i,j)}_r$ determined by the architecture parameters. The meta multigraph is then retrained from scratch to generate node representations for different downstream tasks, \ie, node classification and recommendation.

\subsection{Partial Message based Search Algorithm}
\label{m2} 
DiffMG is 
most related 
to our differentiable multigraph search. In each iteration, DiffMG samples one candidate message passing type on each edge for the forward propagation and backpropagation, and the strongest message passing type is 
most likely to be sampled, resulting in random and insufficient training. DiffMG is efficient and outperforms existing baselines. 
However, one limitation of DiffMG lies in its instability. 
As illustrated in Figure~\ref{fig:seed} of the experiments, DiffMG is only effective in a few random seeds and the performance dramatically declines in most random seeds. 

\begin{figure}[!t]
    \centering
    \includegraphics[width=0.95\linewidth]{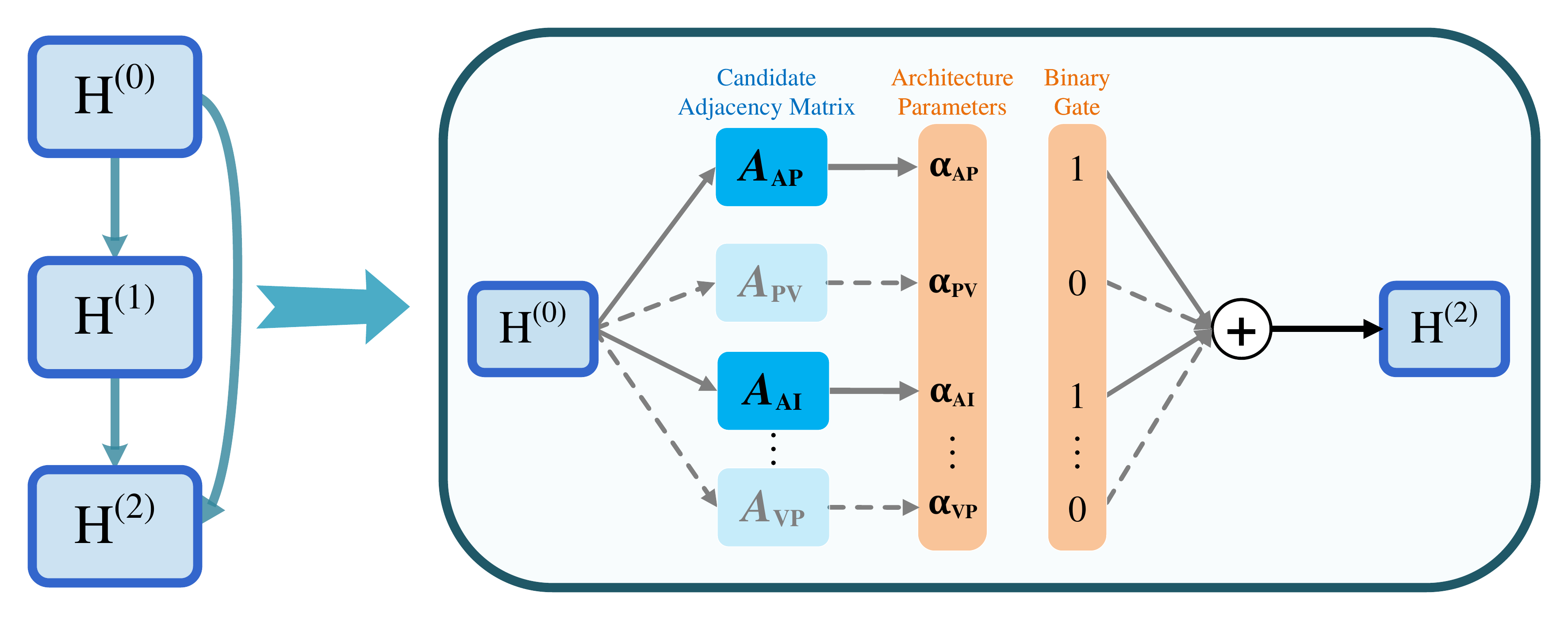}
    \caption{Illustration of the proposed partial message search algorithm. 
    }
    \vspace{-10pt}
    \label{fig:partial}
\end{figure}

To address the instability issue in DiffMG, intuitively we wish to ensure that all message passing types are equally and fully searched in the search stage. An alternative solution is to employ Eq.~\ref{eq:forward}, which trains all possible message passing steps together and formulates the information propagated as a weighted sum over all the paths. However, the architecture parameters of various paths in Eq.~\ref{eq:forward} are deeply coupled and jointly optimized. 
The greedy nature of the differentiable methods inevitably misleads the architecture search due to the deep coupling~\cite{guo2020single}, especially when the number of candidate paths  
is large. It motivates us to propose a new search method for the search stage to achieve stable meta multigraph search as well as overcome the coupling in optimization.

To overcome the coupling optimization,
we define a binary gate $M^{(i,j)}_r$ for each message passing type,  
which assigns $1$ to the selected message passing types and $0$ to the masked ones. 
Specifically, we let paths in each multi-edge be sampled equally and independently, and we set the proportion of $M^{(i,j)}_r=1$ to $1/p$ by regarding $p$ as a hyper-parameter.  
Then we can get the set of all active paths between hyper-node $\bm{H}^{(i)}$ and $\bm{H}^{(j)}$: 
\begin{equation}
S^{(i,j)}=\{r|M^{(i,j)}_r=1,\forall r \in \mathcal{R}^{(i,j)} \}.
\label{eq:sample}
\end{equation}


As illustrated in Figure~\ref{fig:partial}, by introducing the binary gates, only $1/p$ paths of message passing steps are active. 
Then we formulate the information propagated from $\bm{H}^{(i)}$ to $\bm{H}^{(j)}$ as a weighted sum over active candidate message passing steps: 
\begin{align}
\bm{H}^{(j)}=\sum_{i<j}
\sum_{r\in S^{(i,j)}}p^{(i,j)}_r {f}\left(\bm{\mathcal{A}}^{(i,j)}_r,\bm{H}^{(i)}\right),
\label{eq:forward_ours}
\end{align}
where path strength $p^{(i,j)}_r$ is calculated by Eq.~\ref{eq:softmax}.
Since message sampling masks $M^{(i,j)}$ are involved in the computation graph, parameters updated in Eq.~\ref{eq:forward_ours} can be calculated through backpropagation. 
The overall algorithm is given in Algorithm 1.  
\begin{algorithm}[!t]
	\caption{Search algorithm}
	\label{alg:ACA}
	\begin{algorithmic}[1]
		\REQUIRE~~\\
		Network weights $\bm{\omega}$; Architecture parameters $\bm{\alpha}$;  \\
		Number of iterations $T$; Sampling proportion $1/p$.\\
        \ENSURE ~~\\
        Searched architecture parameters $\bm{\alpha}$.
        \STATE Initialize network weights $\bm{\omega}$ and architecture parameters $\bm{\alpha}$
 		\FOR{each iteration $t \in [1,T]$}
\STATE Randomly sample $1/p$ candidate message passing steps in each edge. 
The collection of network weights and architecture parameters of 
sampled paths 
is denoted as $\bar{\bm{\omega}}$ and $\bar{\bm{\alpha}}$, respectively
\STATE Update weights $\bar{\bm{\omega}}$ by  $\nabla_{\bar{\bm{\omega}}}{\cal L}_{train}(\bar{\bm{\omega}}, \bar{\bm{\alpha}})$
\STATE Execute step $3$ again
\STATE Update parameters $\bar{\bm{\alpha}}$ by  $\nabla_{\bar{\bm{\alpha}}}{\cal L}_{val}(\bar{\bm{\omega}}, \bar{\bm{\alpha}})$
\ENDFOR
\RETURN Searched architecture parameters $\bm{\alpha}$.
	\end{algorithmic}
	
\end{algorithm}

\subsection{Meta Multigraph-based Architecture Derivation}
\label{m3}

Once the training of architecture parameters has been completed, we can then derive the compact architecture by pruning redundant paths. As far as we know, all existing differentiable architecture search algorithms for CNNs choose one path with the highest path strength on each edge. As retaining multiple paths means employing multiple types of operations between two node representations, which will damage the performance in most cases.

DiffMG inherits the derivation strategy of these methods to generate meta graphs. However, instead of operations, DiffMG searches the message passing types that determine which messages are propagated between different types of node representations in a meta graph. Considering that an HIN consists of multiple node types and edge types, deriving a single message passing type between two different node representations is insufficient and inflexible to encode rich semantic information. As shown in Figure~\ref{fig:meta-structure} (c), there are multiple paths between $H^{(0)}$ and $H^{(2)}$, which can not be learned by the traditional derivation strategy. Another issue caused by deriving a single path is that some effective message passing types with similar but weaker path strengths will be dropped. So simply selecting the message passing type with the highest path strength may reject potentially good architectures.

To address the above issues, we propose to derive a meta multigraph as a heterogeneous message passing layer, making the message passing types more diverse than simply retaining the one with the highest path strength.  
An alternative solution to derive a meta multigraph is to set a threshold $\tau$, message passing types with path strengths above which are retained. As path strengths are keeping changing during the search stage, $\tau$ needs to be changed with them. Here, we set the threshold $\tau^{(i,j)} $ as a value between the largest and the smallest path strengths in each multi-edge $\mathcal{R}^{(i,j)}$: 
\begin{equation}
\tau^{(i,j)}= \lambda \cdot
\max\limits_{r \in \mathcal{R}^{(i,j)}}\{p^{(i,j)}_r\}+ (1-\lambda) \cdot \min\limits_{r \in \mathcal{R}^{(i,j)}}\{p^{(i,j)}_r\},
\label{eq:tau}
\end{equation}
where $p^{(i,j)}_r$ is inherited from the search stage, $\lambda \in [0,1]$ is a hyper-parameter controlling the number of retrained paths in each multi-edge.  

Then, we formulate the information propagated from $\bm{H}^{(i)}$ to $\bm{H}^{(j)}$ in the derived meta multigraph as an unweighted sum over candidate paths with path strengths above $\tau^{(i,j)}$:
\begin{equation}
\bm{H}^{(j)}=\sum_{i<j}
\sum_{r\in \hat{S}^{(i,j)}}{f}\left(\bm{\mathcal{A}}^{(i,j)}_r,\bm{H}^{(i)}\right),
\label{eval}
\end{equation}
\begin{equation}
\hat{S}^{(i,j)}=\{r|p^{(i,j)}_r \ge \tau^{(i,j)},\forall r \in \mathcal{R}^{(i,j)}\},
\label{handcrafted}
\end{equation}
where $\hat{S}^{(i,j)}$ is the set of all retained paths between hyper-node $\bm{H}^{(i)}$ and $\bm{H}^{(j)}$. Then, the derived meta multigraph can be used as the target-net for retraining from scratch.

\subsection{Differences to Prior Works}
\label{m4}
\begin{table}[t]
	\centering
	\caption{Comparison of PMMM with related NAS algorithms. \textit{Coupling} denotes whether all architecture parameters are coupling optimized. \textit{Probability} denotes the updating probability of each path in one iteration.
		}
	\label{tab:search_and_derive}
	\begin{threeparttable}[b]
	\small
	\resizebox{0.40\textwidth}{!}{
	\begin{tabular}{l  c c c}
		\toprule
		\multirow{2}{*}{Method}       & 
		\multicolumn{2}{c}{Search}  
		 &  \multirow{2}{*}{Derivation}    \\ 
		&Coupling&Probability&  \\
		\midrule
		DARTS
		& \ding{51}   &      high       &    single path        \\
		PC-DARTS
		&  \ding{51}  &      high       &    single path        \\
		ProxylessNAS
		&  \ding{55}  &      low       &    single path        \\
		SPOS
		&  \ding{55}   &     low        &   single path         \\
		DiffMG
		&   \ding{55}    &   low          &   single path         \\
		PMMM       & \ding{55}    &   middle          &   multiple paths         \\ \bottomrule
	\end{tabular}}
	\end{threeparttable}
\end{table}

Table~\ref{tab:search_and_derive} details the differences between our approach and related differentiable NAS algorithms, including DARTS~\cite{liu2018darts}, PC-DARTS~\cite{xu2019pc}, ProxylessNAS~\cite{cai2018proxylessnas}, SPOS~\cite{guo2020single}, and DiffMG~\cite{yao2019differentiable}. 
The first four algorithms search convolution or pooling operations in CNNs instead of meta-structures in GNNs. We ignore these differences and focus on the algorithms.
In contrast to DARTS and PC-DARTS, our search algorithm does not jointly optimize all the architecture parameters, which reduces the inconsistency between the search and evaluation phase. Compared to ProxylessNAS, SPOS, and DiffMG, our search algorithm updates each message passing step with a higher probability,  
avoiding unfairness caused by insufficient training. Regarding the derivation strategy, our approach is distinct from all the above methods. \par

\section{Experiments}
\label{sec:experiments}

\begin{table*}[!t]
    \small
	\centering
	\caption{Macro-F1 ($\%$) and Micro-F1 ($\%$) on the node classification task (mean in percentage ± standard deviation). The best and second best results are shown in \textbf{bold} and \underline{underline}.}
	\label{tab:nc}
    \begin{threeparttable}[b]
	\small
	\resizebox{0.88\textwidth}{!}{
	\begin{tabular}{>{\rowmac}l>{\rowmac}c>{\rowmac}c>{\rowmac}c>{\rowmac}c>{\rowmac}c>{\rowmac}c}
		\toprule
		\multirow{2}{*}{Architecture}& \multicolumn{2}{c}{DBLP} & \multicolumn{2}{c}{ACM} & \multicolumn{2}{c}{IMDB} \\
		& Macro-F1 & Micro-F1 & Macro-F1 & Micro-F1 & Macro-F1 & Micro-F1\\\midrule
		metapath2vec  & 89.93$\pm$0.45 & 88.62$\pm$0.44 & 67.13$\pm$0.50 & 68.24$\pm$0.41 & 40.82$\pm$1.48 & 43.47$\pm$1.21   \\  
		GCN    & 90.46$\pm$0.41 & 91.47$\pm$0.34 & 92.56$\pm$0.20  & 91.22$\pm$0.64 & 55.19$\pm$0.99 & 54.63$\pm$1.03\\
        GAT    & 93.92$\pm$0.28 & 93.39$\pm$0.30 & 92.50$\pm$0.23  & 92.86$\pm$0.29 & 53.37$\pm$1.27 & 53.64$\pm$1.43 \\
        HAN    & 92.13$\pm$0.26 & 92.05$\pm$0.62 & 91.20$\pm$0.25  & 91.57$\pm$0.38 & 55.09$\pm$0.67 & 55.32$\pm$0.81 \\
        MAGNN  & 92.81$\pm$0.30 & 92.58$\pm$0.21 & 91.15$\pm$0.19  & 91.06$\pm$0.33 & 56.44$\pm$0.63 & 56.60$\pm$0.72 \\
        GTN    & 93.98$\pm$0.32 & 94.01$\pm$0.28 & 92.62$\pm$0.17  & 91.78$\pm$0.59 & 59.68$\pm$0.72 & 59.40$\pm$1.44 \\
        HGT    & 93.67$\pm$0.22 & 93.49$\pm$0.25 & 91.83$\pm$0.23  & 91.12$\pm$0.23 & 59.35$\pm$0.79 & 59.42$\pm$0.74 \\
        GraphMSE   & 94.08$\pm$0.14 & 94.44$\pm$0.13 & 92.58$\pm$0.50  & 92.54$\pm$0.49 & 57.60$\pm$2.13 & 62.37$\pm$1.03 \\
        DiffMG & 94.45$\pm$0.15 & 94.85$\pm$0.27 & 92.65$\pm$0.15  & 92.21$\pm$0.31 & 61.04$\pm$0.56 & 61.81$\pm$1.59  \\
        \midrule
        Multigraph (ours) &  \textbf{94.75$\pm$0.11} & \underline{95.38$\pm$0.13} & \textbf{93.36$\pm$0.09} & \textbf{93.27$\pm$0.09} & \underline{61.46$\pm$0.77}  & \underline{63.44$\pm$1.01} \\
        \model (ours) &  \underline{94.69$\pm$0.10} & \textbf{95.40$\pm$0.10} & \underline{92.76$\pm$0.14} & \underline{92.65$\pm$ 0.12} & \textbf{61.69$\pm$0.40} & \textbf{64.31$\pm$0.77} \\
		\bottomrule
	\end{tabular}}
	\end{threeparttable}
\end{table*}

For experiments, we first compare PMMM with baselines on two representative tasks to evaluate its performance.  
We then evaluate the efficiency of PMMM and visualize our searched architectures to analyze how meta multigraph improves the performance.
We also compare PMMM with differentiable meta graph search to show its stability under various
settings. 
In the end, we conduct parameter analysis on hyper-parameter $\lambda$ to show the flexibility of PMMM. 

\subsection{Experimental Setup}
\label{sec:setup}
\subsubsection{Datasets}
We evaluate our method on two popular tasks~\cite{yang2020heterogeneous}: node classification and recommendation. The goal of the node classification task is to predict the correct labels for nodes based on network structure and node features. We use three widely-used real-world datasets
: DBLP, ACM, and IMDB. Regarding the recommendation task, we aim to predict links between source nodes (\eg, users) and target nodes (\eg, items). We adopt three widely used heterogeneous recommendation datasets
: Amazon, Yelp, and Douban Movie (abbreviated as Douban). The details of all datasets are shown in Appendix.

\subsubsection{Baselines}
We compare our method with eleven methods, including:  
1) a random walk based network embedding method,  metapath2vec~\cite{dong2017metapath2vec};  
2) two homogeneous GNNs, \ie, GCN~\cite{kipf2016semi} and GAT~\cite{velivckovic2017graph};  
3) five heterogeneous GNNs, \ie, HAN~\cite{wang2019heterogeneous}, MAGNN~\cite{fu2020magnn}, 
GTN~\cite{NIPS2019_9367}, HGT~\cite{hu2020heterogeneous}, and GraphMSE~\cite{li2021graphmse};  
4) three AutoML methods, \ie, GEMS~\cite{han2020genetic} for recommendation, HGNAS~\cite{gao2021heterogeneous} for node classification, and DiffMG~\cite{DBLP:conf/kdd/DingYZZ21}. 
Specially, HGNAS employs a different data division and does not provide the source code, so we  
do a separate comparison with HGNAS using its data division in Appendix.
More details of these methods and the differences with our model can be found in Appendix.

\subsubsection{Parameter Settings} 
Following DiffMG, we run the search algorithm three times with different random search seeds 
to derive the meta multigraph from the run that achieves the best validation performance. For a fair comparison, the candidate message passing types are the same with DiffMG and the searching epochs are set to $30$. We set $p = 2$ for most datasets, \ie, only $1/2$ paths are randomly sampled on each edge, except for DBLP and ACM with small $K$, we set $p = 1$. 
To coordinate with baselines, we set the steps $N = 4$, which is the same as the length of meta-paths learned by GTN and meta graph searched by DiffMG. 
Besides, we set $\lambda = 0.9$. 
To demonstrate the effectiveness of our meta multigraph strategy, we combine the search algorithm of DiffMG and our meta multigraph derivation strategy to obtain new architectures, which we call Multigraph, and compare them with DiffMG. 
We put the settings for baselines in Appendix.
Experiments are conducted on a single RTX $2080$ Ti GPU with 11GB memory.  

\subsubsection{Evaluation Metrics}
For evaluation, we use Macro-F1 score as well as Micro-F1 score for the node classification task and AUC (area under the ROC curve) for the recommendation task. For each method, we repeat the process for 10 runs with different random training seeds and report the average score and standard deviation.


\subsection{Comparison on Node Classification}
\label{sec:nc}
Table~\ref{tab:nc} summarizes the results of the proposed model and the baselines.   
First, Heterogeneous GNNs relying on manually designed meta-paths, such as HAN and MAGNN, do not achieve desirable performance. They can even perform worse than homogeneous GNNs like GAT, which suggests that hand-crafted rules may even have adverse implications.
Second, DiffMG outperforms heterogeneous GNNs employing meta-path, demonstrating the power of NAS as well as the advantages of meta graphs over meta-paths. However, the performance of DiffMG is fragile as it relies on carefully chosen random seeds and meta graph lengths as shown in the subsequent Figure~\ref{fig:seed} and Figure~\ref{fig:step}. 
Finally, PMMM outperforms all the state-of-the-art models consistently, which further indicates the importance of automatically leveraging task-dependent semantic information in HINs. Moreover, the performance improvements of Multigraph over DiffMG validate that a meta multigraph surpasses a meta graph in the representation learning on HINs. Note that the meta multigraph strategy can not address the issue of instability. So the performance of Multigraph is worse than PMMM in most other search seeds.

\begin{table}[tb]
	\centering
	\caption{AUC ($\%$) on the recommendation task (mean in percentage ± standard deviation). }
	\label{tab:lp}
	\begin{threeparttable}[b]
	\small
	\resizebox{0.47\textwidth}{!}{
	\begin{tabular}{>{\rowmac}l>{\rowmac}c>{\rowmac}c>{\rowmac}c}
		\toprule
		Architecture & Amazon & Yelp & Douban \\
		\midrule
		metapath2vec  &  58.17$\pm$0.14 & 51.98$\pm$0.14    & 51.60$\pm$0.07    \\  
		GCN    & 66.64$\pm$1.00 & 58.98$\pm$0.52  & 77.95$\pm$0.05  \\
        GAT    & 55.70$\pm$1.13   & 56.55$\pm$0.05    & 77.58$\pm$0.33  \\
        HAN    & 67.35$\pm$0.11   & 64.28$\pm$0.20    & 82.65$\pm$0.08 \\
        MAGNN  & 68.26$\pm$0.09   & 64.73$\pm$0.24    & 82.44$\pm$0.17  \\
        GEMS  &  70.66$\pm$0.14   & 65.12$\pm$0.27    & 83.00$\pm$0.05 \\
        GTN    & 71.82$\pm$0.18   & 66.27$\pm$0.31    & 83.26$\pm$0.10 \\
        HGT    & 74.75$\pm$0.08 & 68.07$\pm$0.35    & 83.38$\pm$0.06    \\
        DiffMG & 75.28$\pm$0.08 & 68.77$\pm$0.13    & 83.78$\pm$0.09    \\
        \midrule
        Multigraph (ours)&  \underline{75.32$\pm$0.04} & \textbf{69.48$\pm$0.13}   & \textbf{83.90$\pm$0.02}  \\
        PMMM (ours)    & \textbf{75.48$\pm$0.04} & \underline{69.27$\pm$0.13}  & \underline{83.88$\pm$0.02} \\
		
		\bottomrule
	\end{tabular}}
	\end{threeparttable}
\end{table}

\subsection{Comparison on Recommendation}
\label{sec:lr}
Table~\ref{tab:lp} reports the AUC results on the recommendation task of our methods and the baselines. 
Both our Multigraph and PMMM surpass the state-of-the-art models on all the three datasets, demonstrating that a meta multigraph surpasses a meta-path or meta graph in capturing the semantic information on HINs. PMMM also outperforms HGT~\cite{hu2020heterogeneous}, a far more complex architecture.
The results indicate the superiority of the NAS method in HINs.

\begin{table}[!t]
	\centering
	\caption{Search cost compared with automated machine learning algorithms on HINs, measured in GPU minutes.}
	\label{tab:search_cost}
	\begin{threeparttable}[b]
    \footnotesize
    \resizebox{0.38\textwidth}{!}{
	\begin{tabular}{c|ccc}
		\hline 
		Method & \   Amazon \ \            & \ \  Yelp  \ \                    & \ Douban  \    \\ \hline  
		GEMS          & 800                 & 1500                      & 2000              \\ 
        DiffMG     &0.7        &1.1                &1.6          \\  
		PMMM       & 0.9  & 1.4  & 2.1 \\ 
        \hline  
	\end{tabular}}
	\end{threeparttable}
\end{table}
\subsection{Efficiency}
Table~\ref{tab:search_cost} compares the search cost of our method against existing works employing NAS for heterogeneous GNNs on three recommendation datasets. HGNAS is omitted as it does not provide the search cost and source code. 
GEMS is time-consuming as it evaluates each candidate model individually. 
For DiffMG and PMMM, we present the overall search cost of three runs.
Both methods combine candidate models into a super-net. So the search cost can be reduced by about $1000$ times, enabling them to be applied to large-scale HINs. 
Compared with DiffMG, our method has a similar search efficiency but exhibits better performance and much-improved stability. The evaluation efficiency is illustrated in Appendix, which has a similar performance to search efficiency.



\begin{figure}[!t]
\centering
\subfigure[DBLP]{\includegraphics[width=0.28\linewidth]{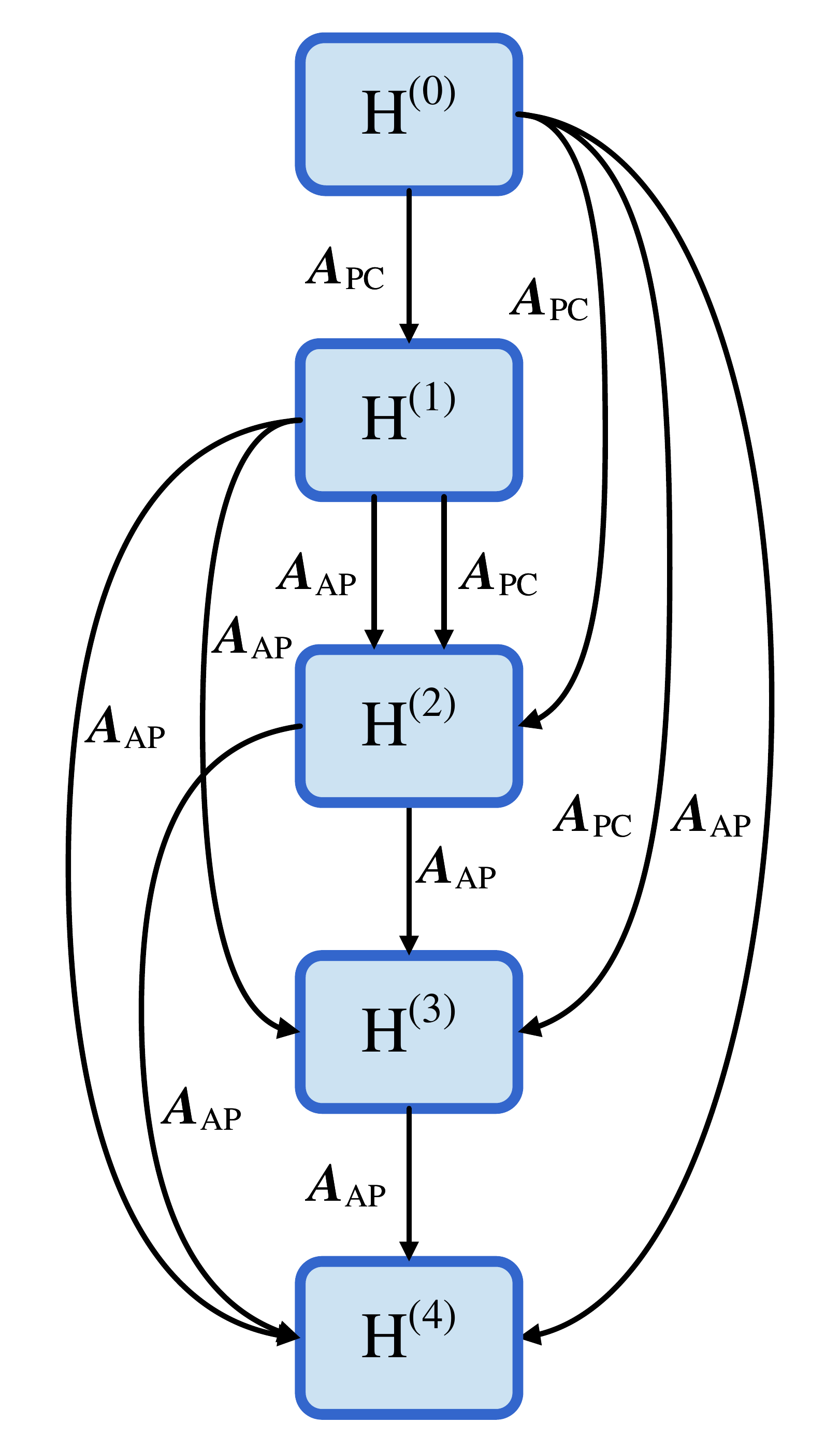}}
\hspace{0.5em}
\subfigure[Amazon]{\includegraphics[width=0.54\linewidth]{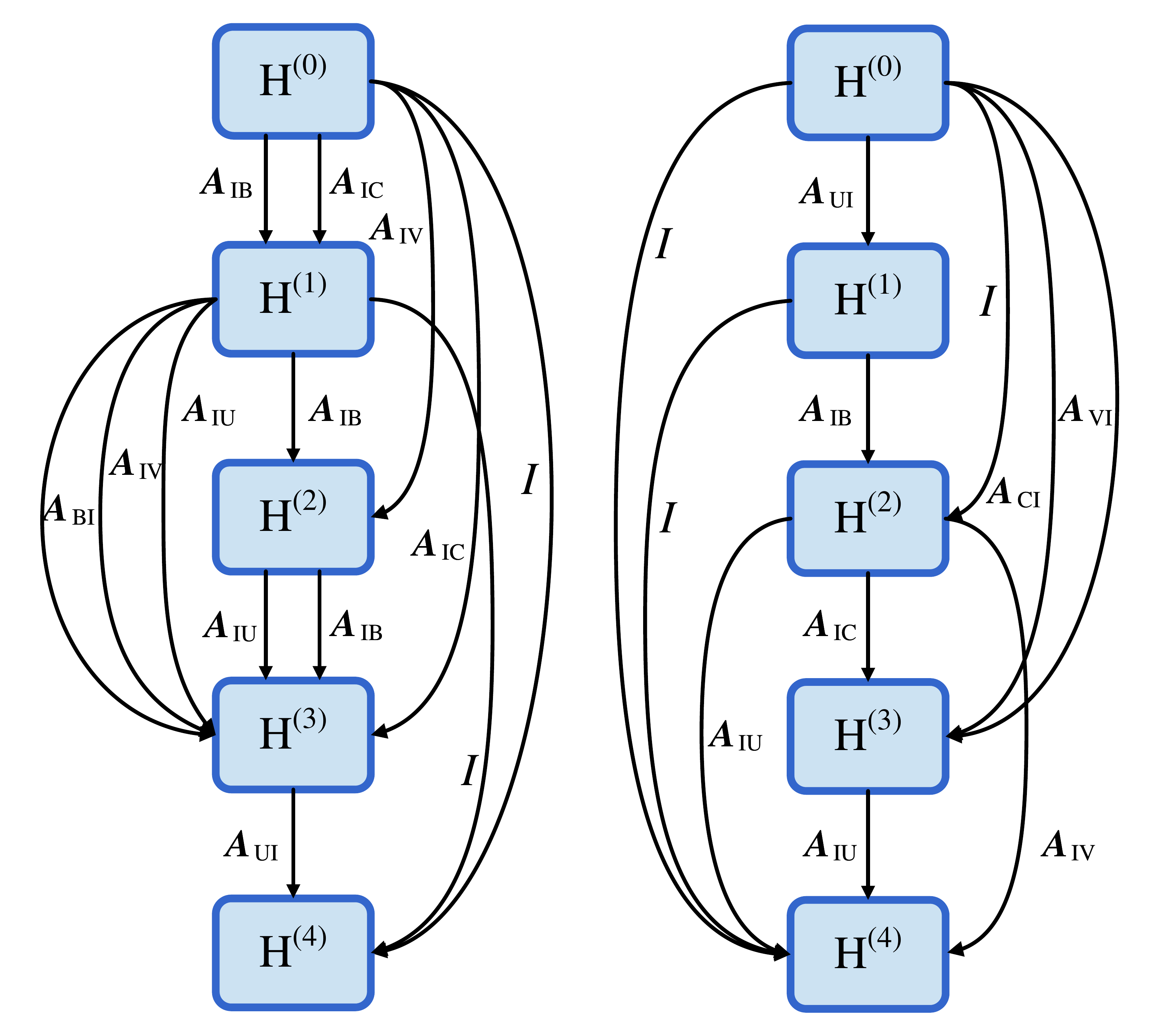}}

\caption{Architectures searched by PMMM. Each edge type is associated with aggregation function, so the searched meta-structures are the final architecture.}
\vspace{-5pt}
\label{fig:visual}
\end{figure}

\begin{figure*}[!t]
\centering
\subfigure[DBLP]{\includegraphics[width=0.16\linewidth]{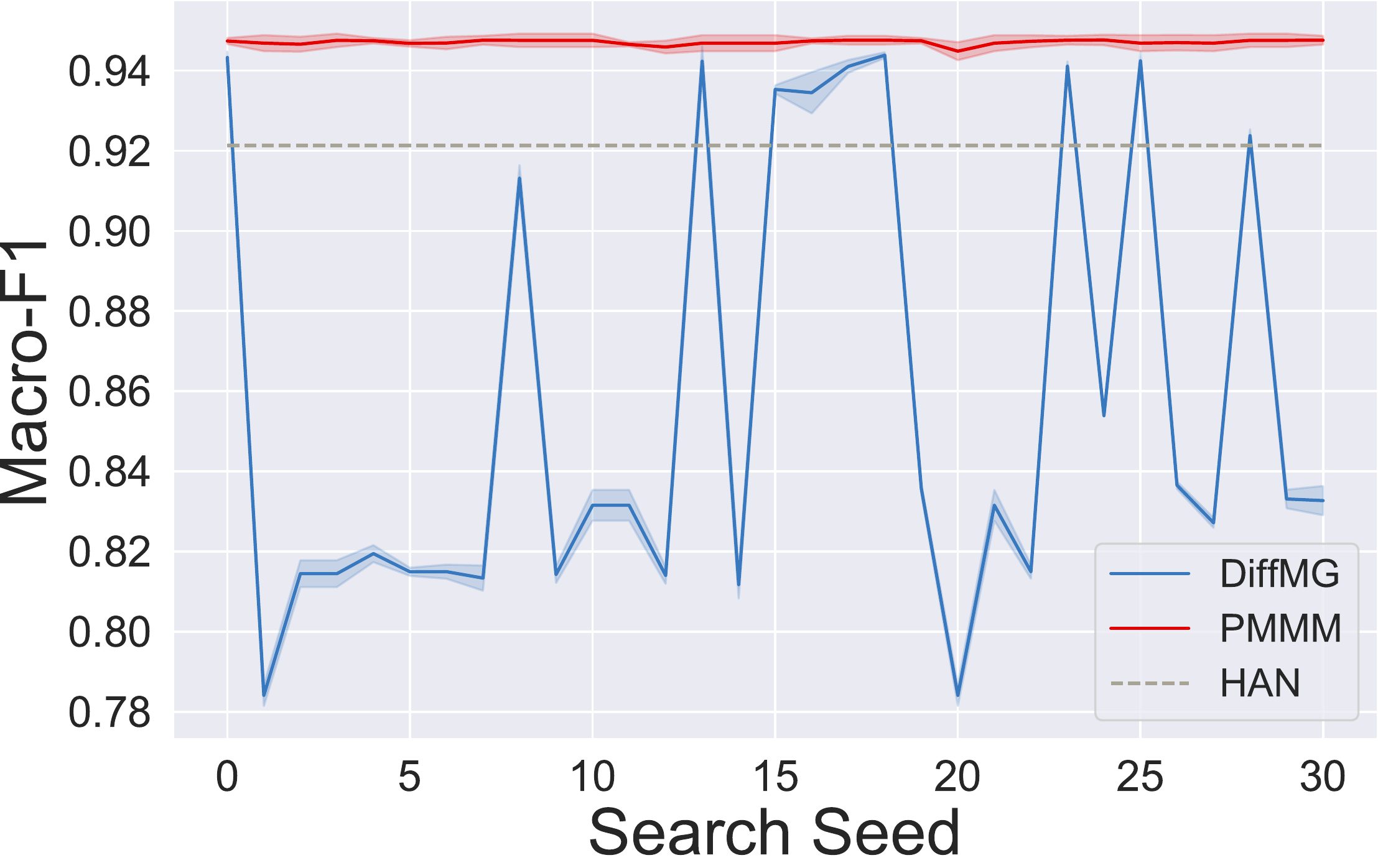}}
\subfigure[ACM]{\includegraphics[width=0.16\linewidth]{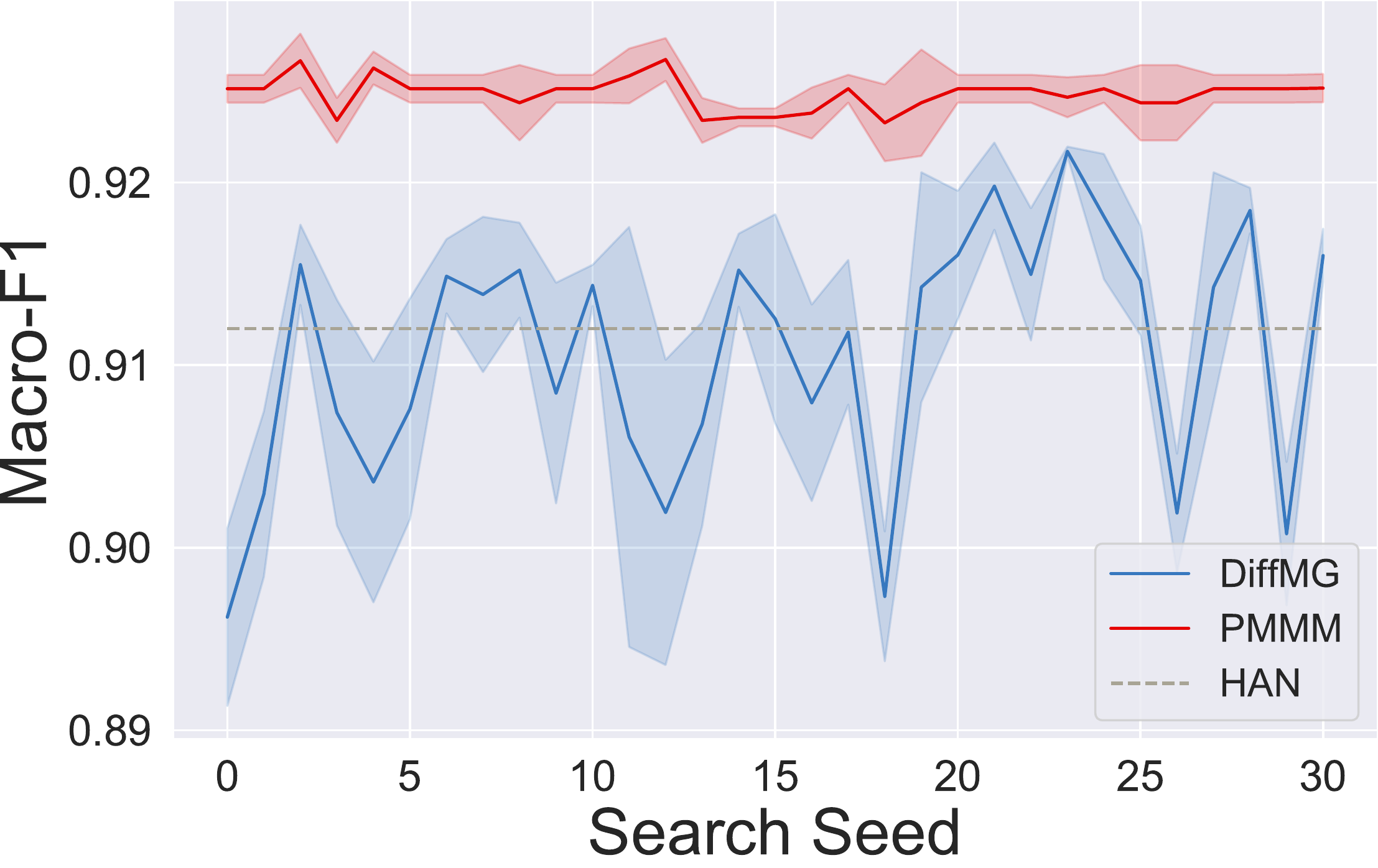}}
\subfigure[IMDB]{\includegraphics[width=0.16\linewidth]{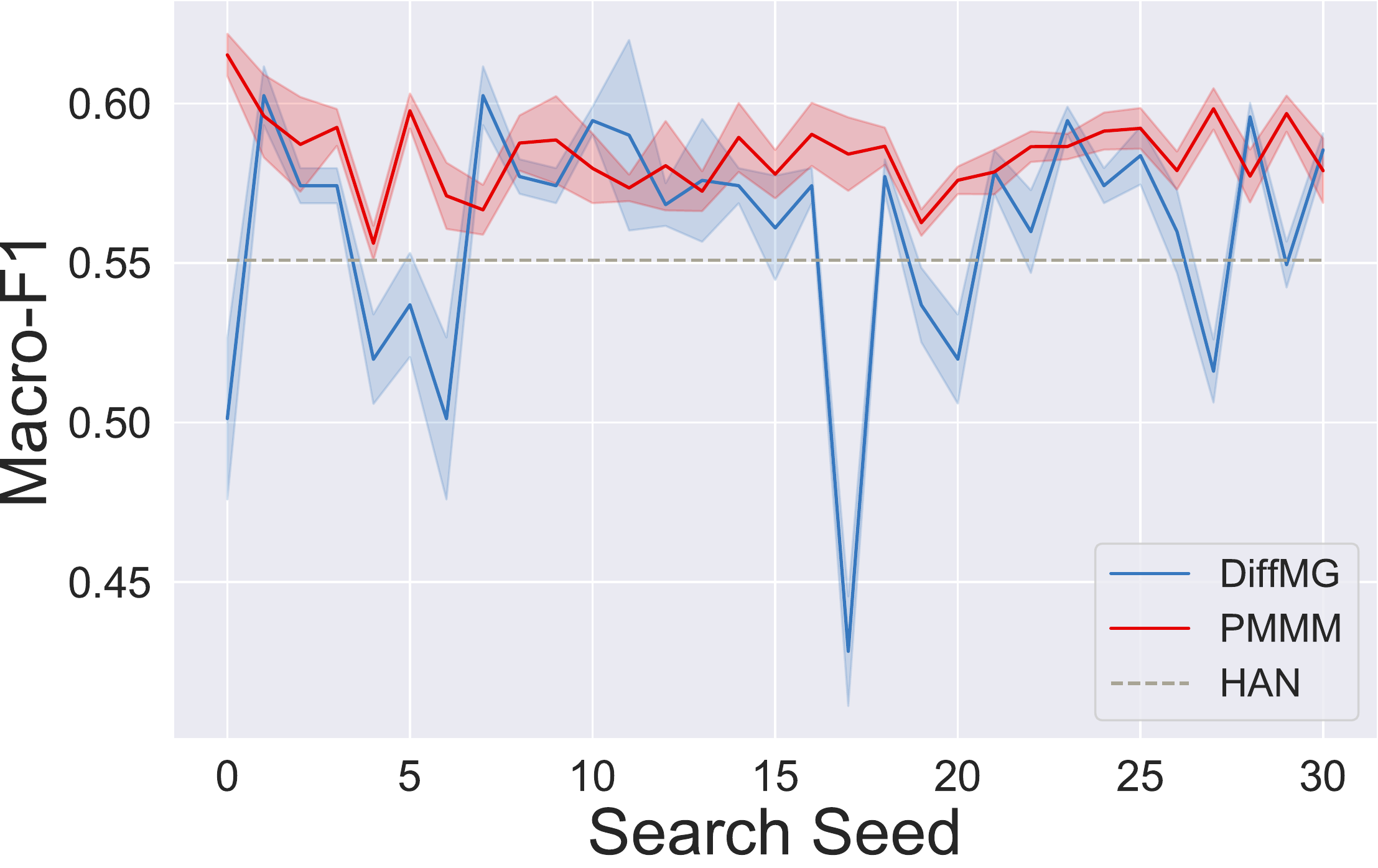}}
\subfigure[Amazon]{\includegraphics[width=0.16\linewidth]{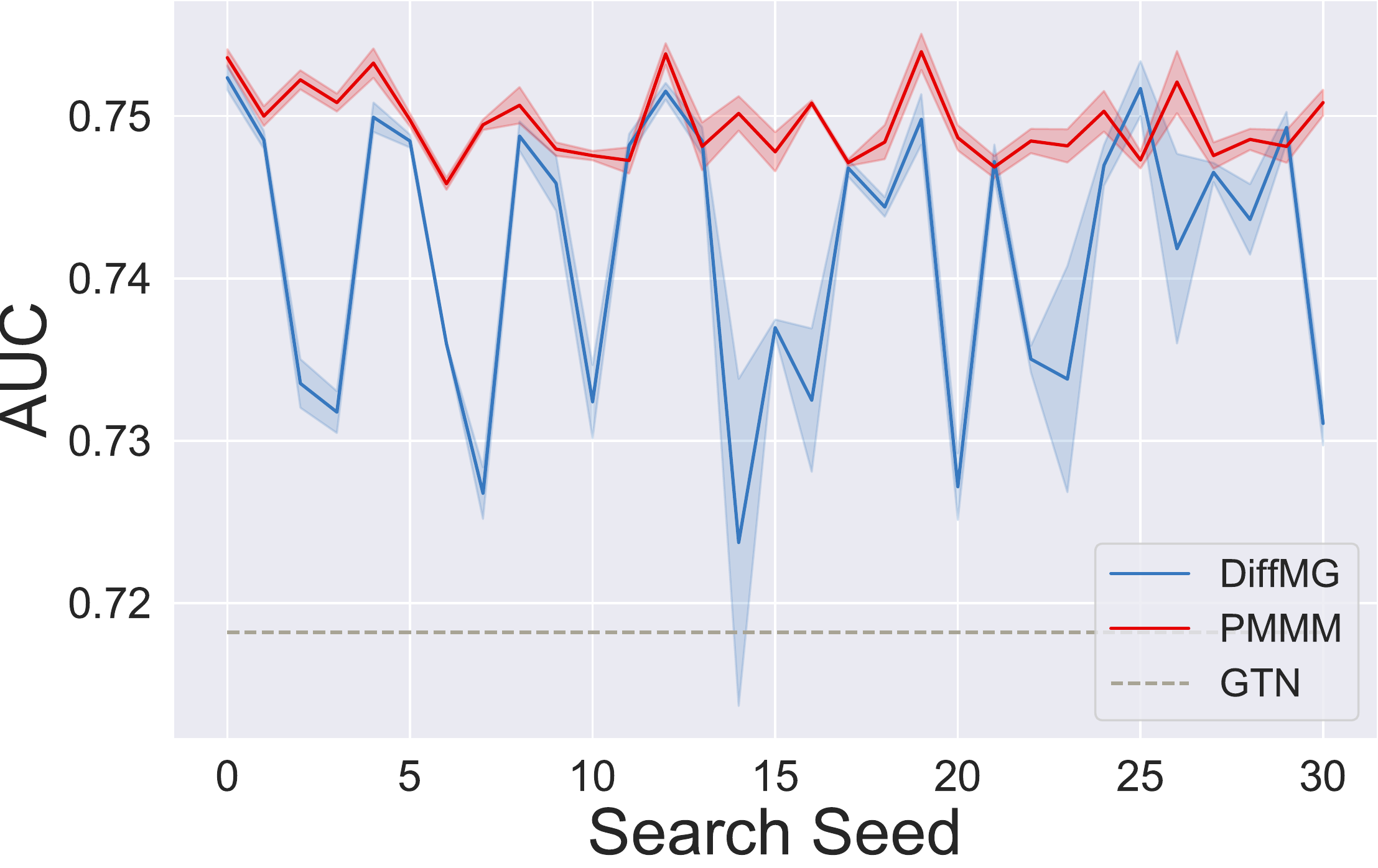}}
\subfigure[Yelp]{\includegraphics[width=0.16\linewidth]{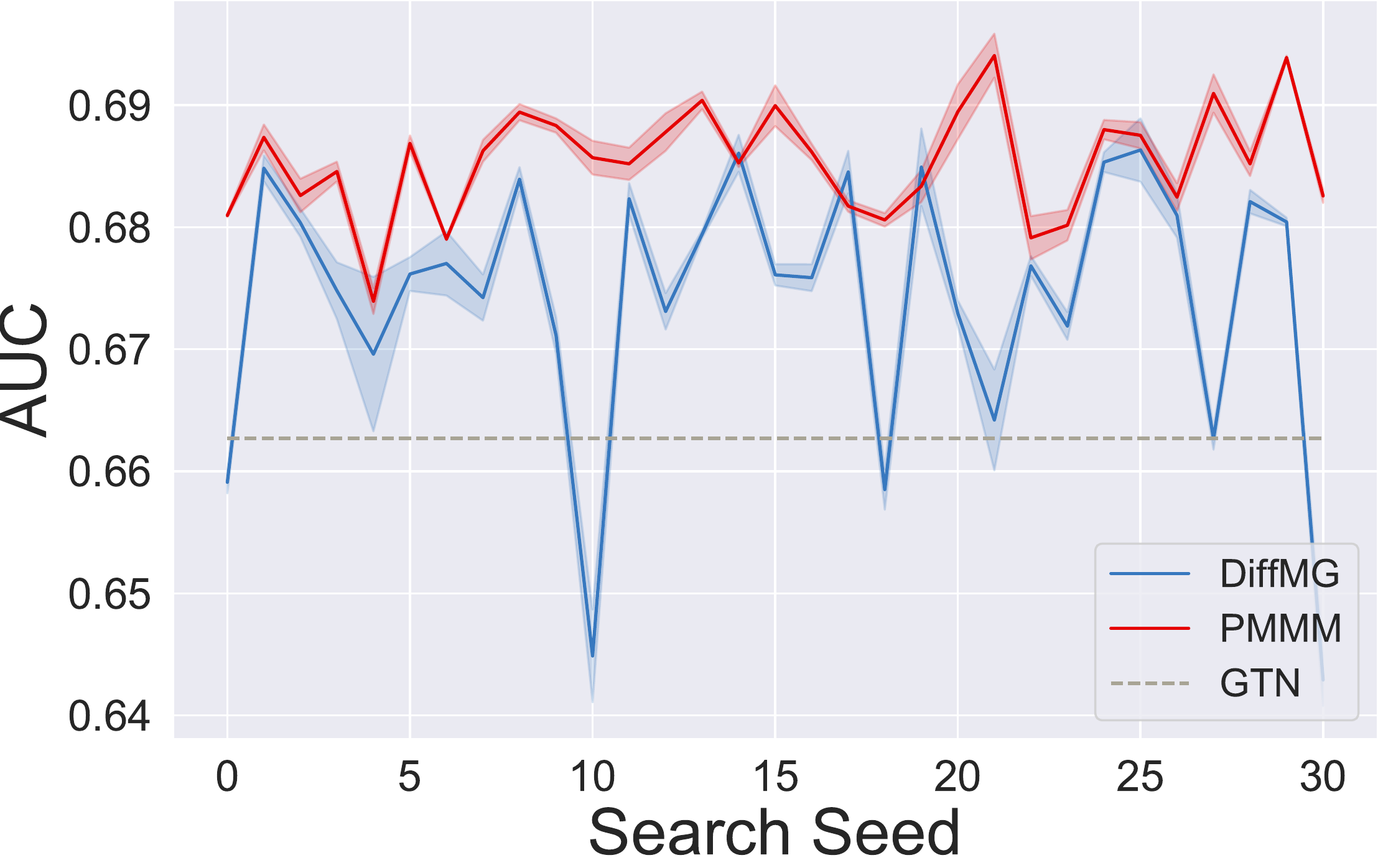}}
\subfigure[Douban]{\includegraphics[width=0.16\linewidth]{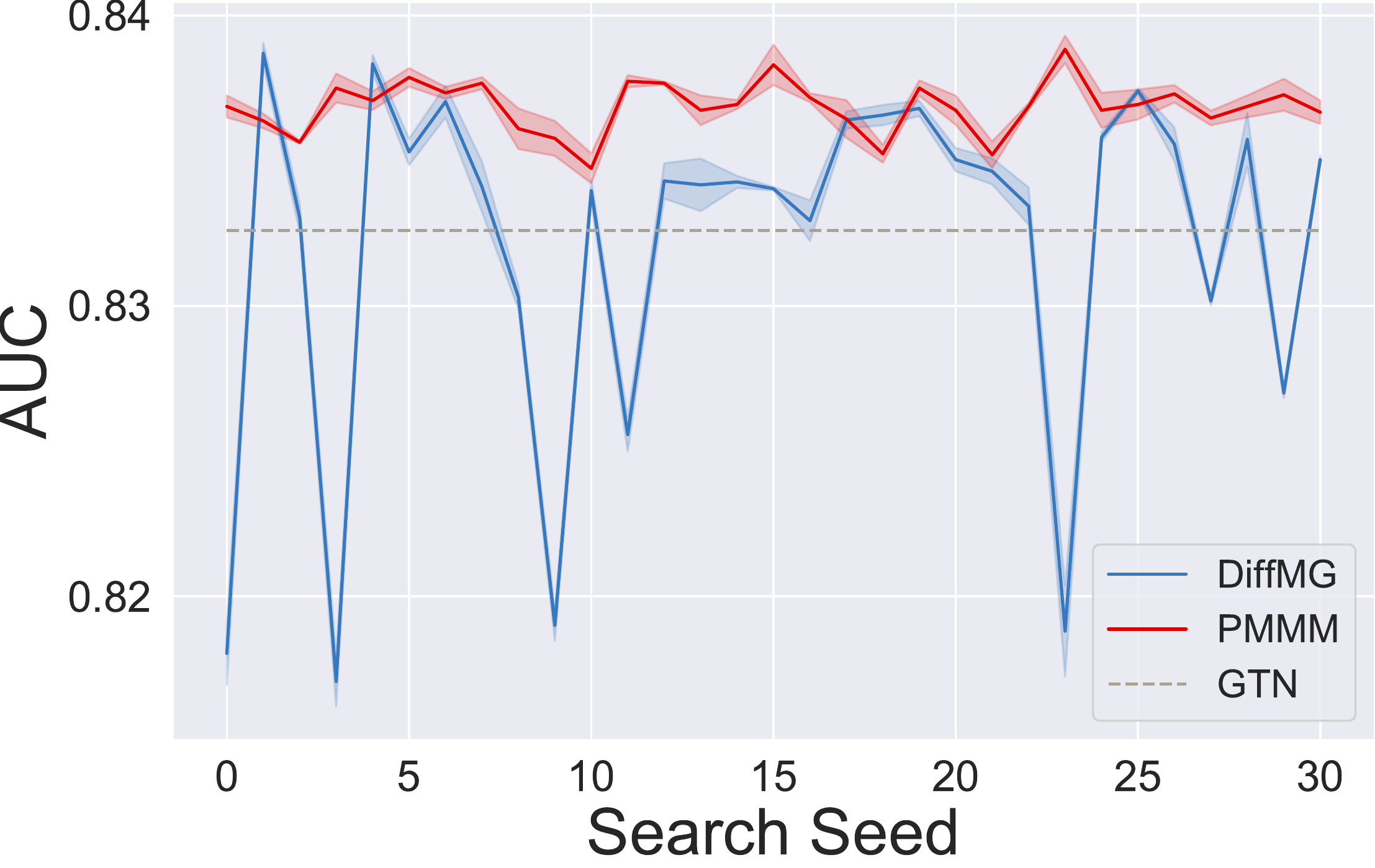}}
\vspace{-5pt}
\caption{Performance on different search seeds (best viewed in color). The first three figures are results for node classification task, followed by three figures for recommendation task. 
}
\label{fig:seed}
\end{figure*}

\begin{figure*}[!t]
\centering
\subfigure[DBLP]{\includegraphics[width=0.16\linewidth]{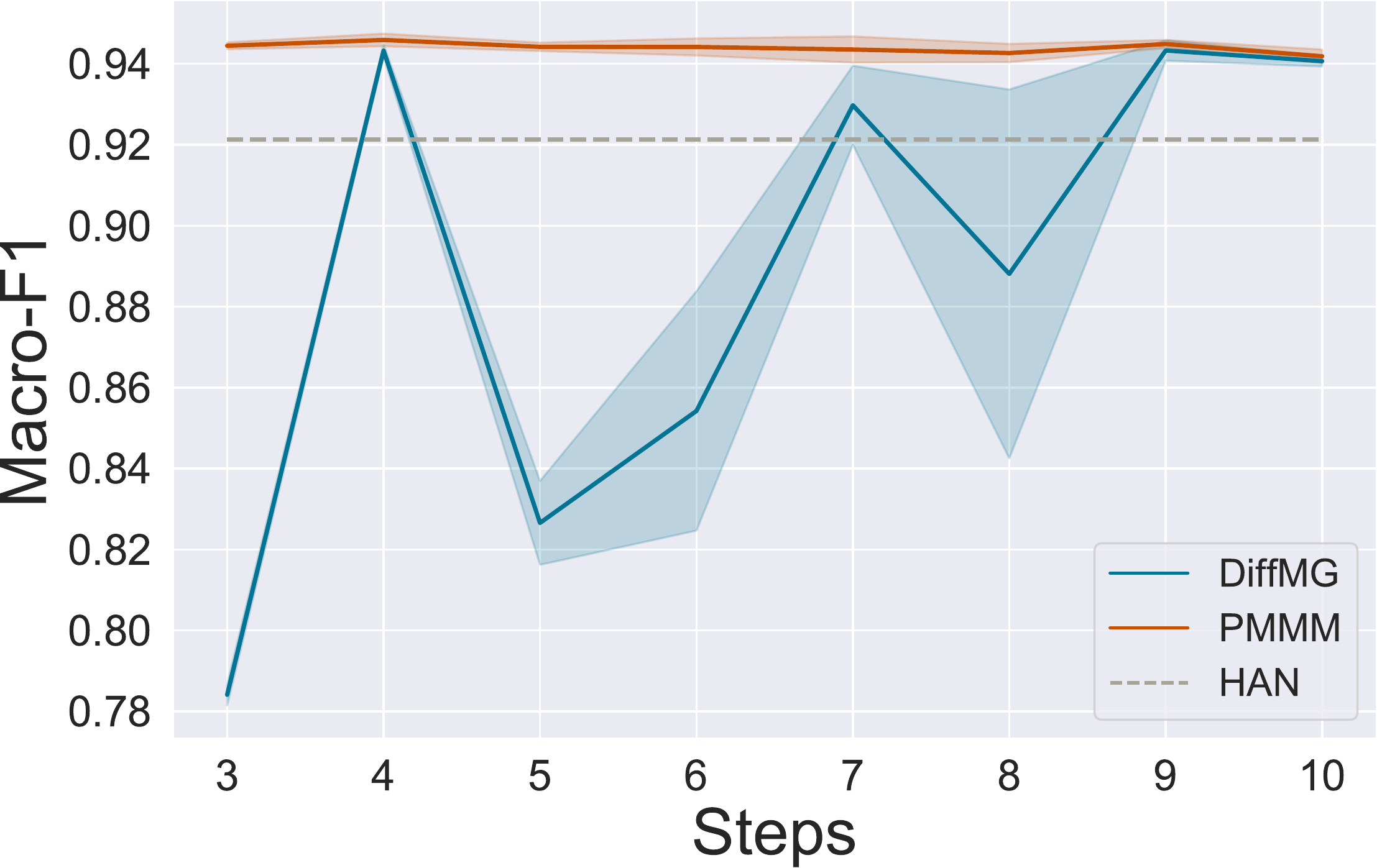}}
\subfigure[ACM]{\includegraphics[width=0.16\linewidth]{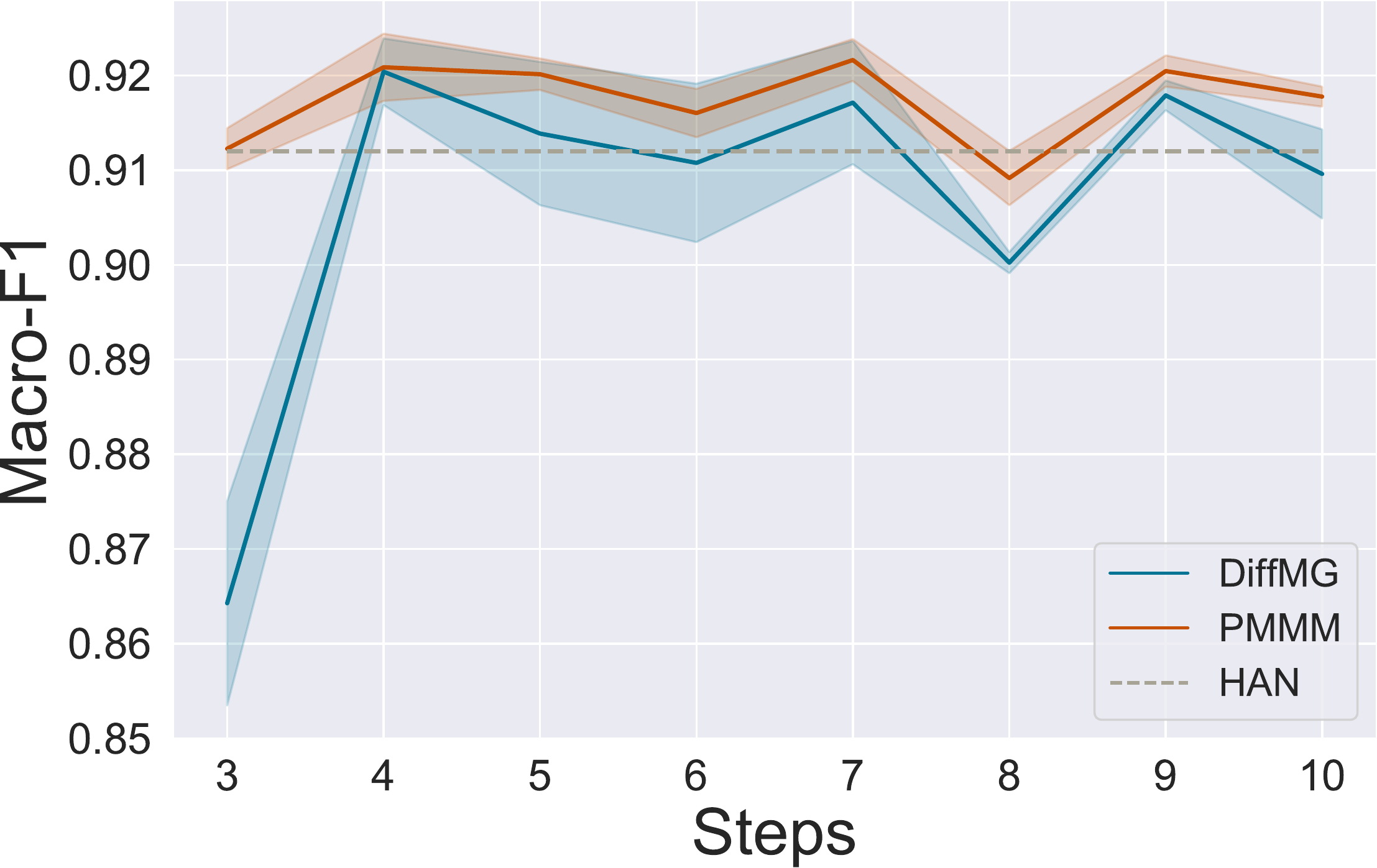}}
\subfigure[IMDB]{\includegraphics[width=0.16\linewidth]{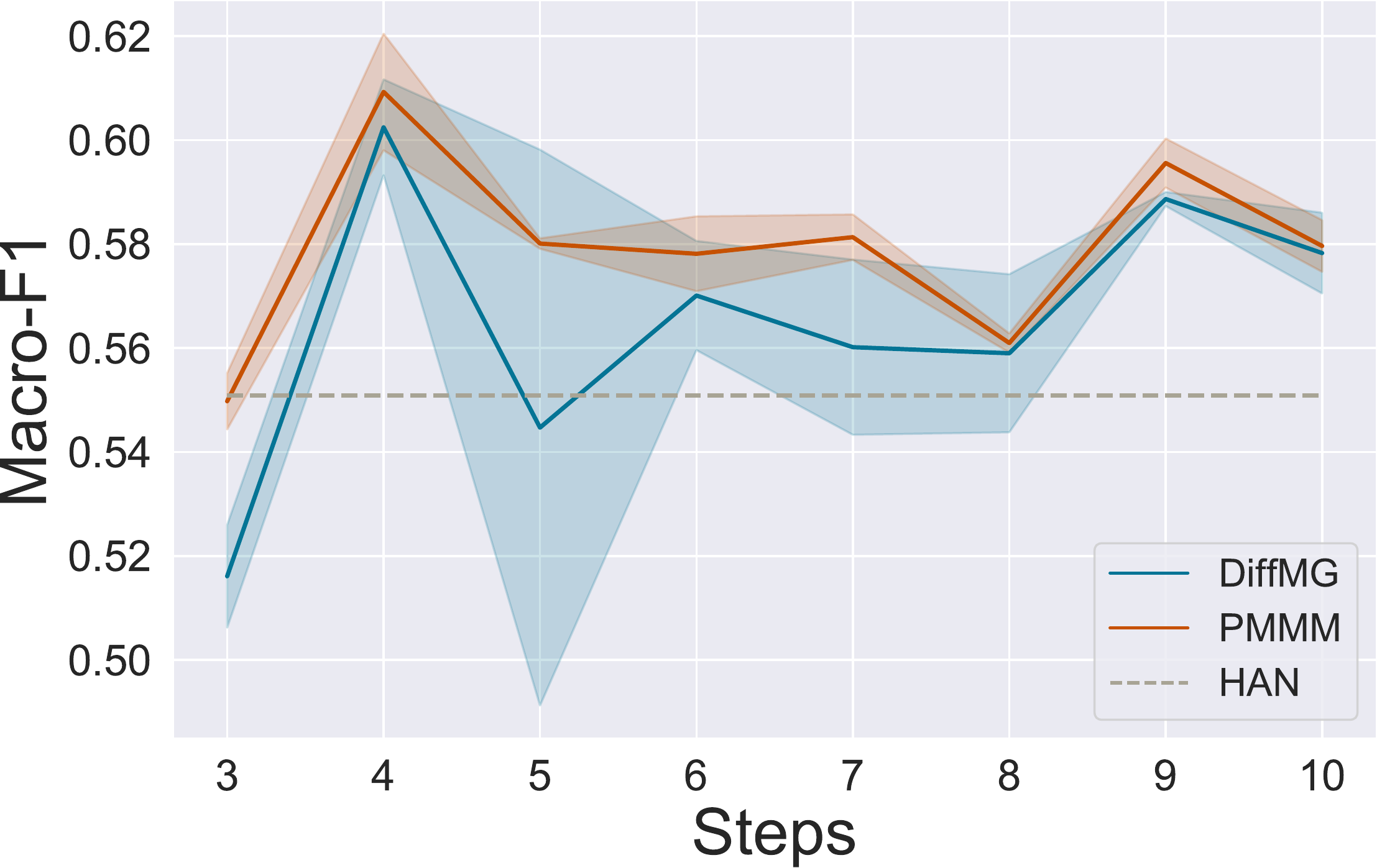}}
\subfigure[Amazon]{\includegraphics[width=0.16\linewidth]{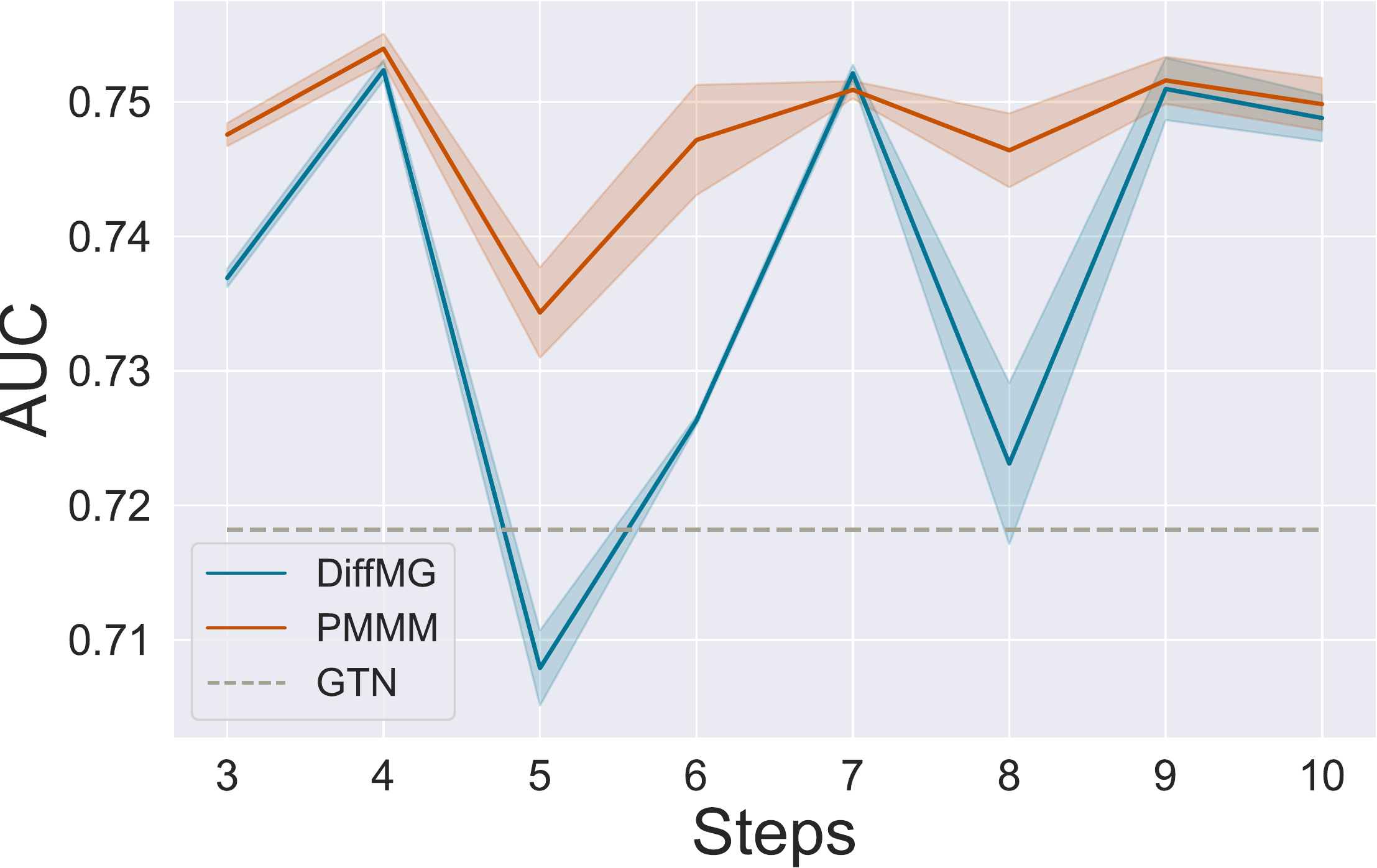}}
\subfigure[Yelp]{\includegraphics[width=0.16\linewidth]{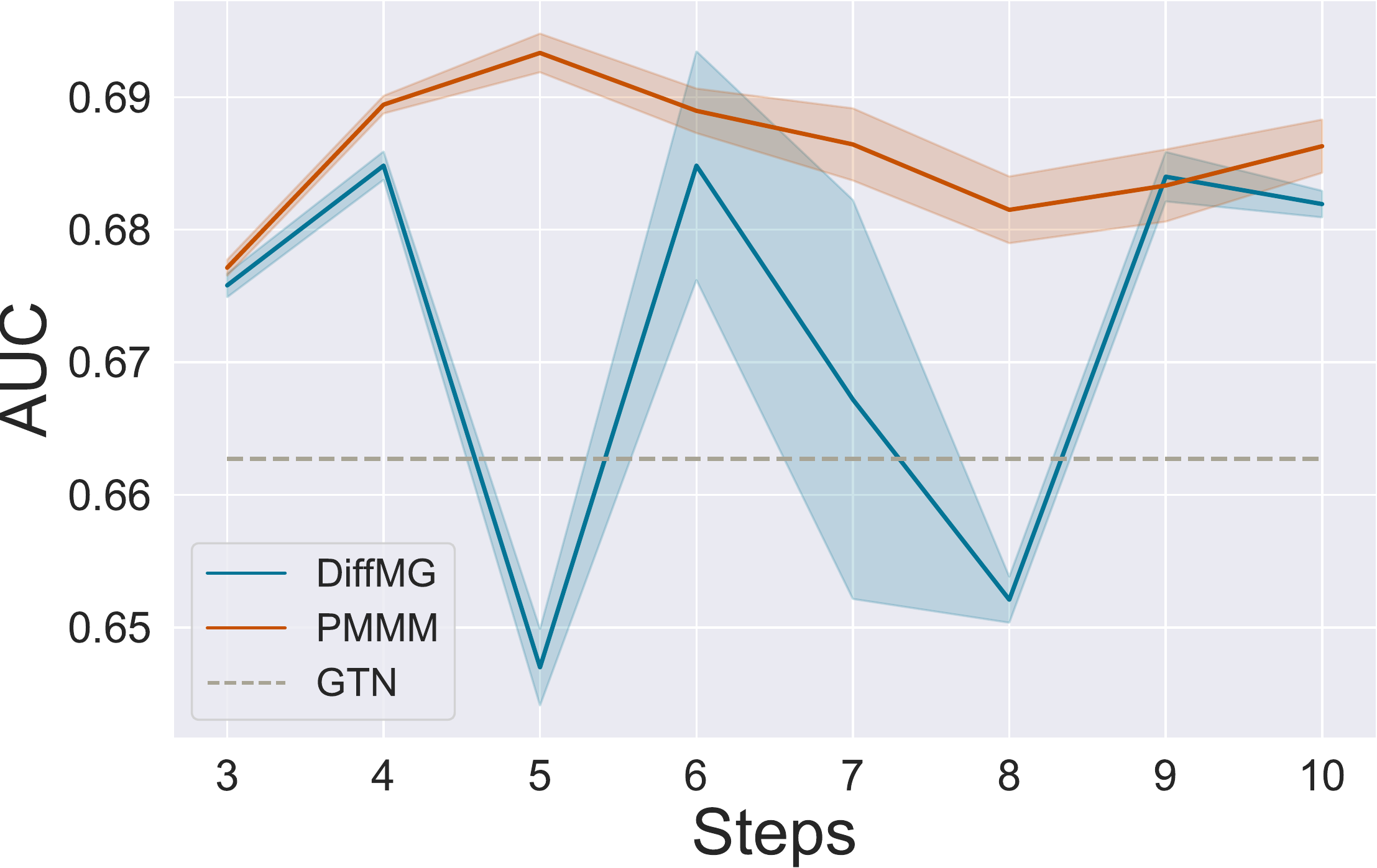}}
\subfigure[Douban]{\includegraphics[width=0.16\linewidth]{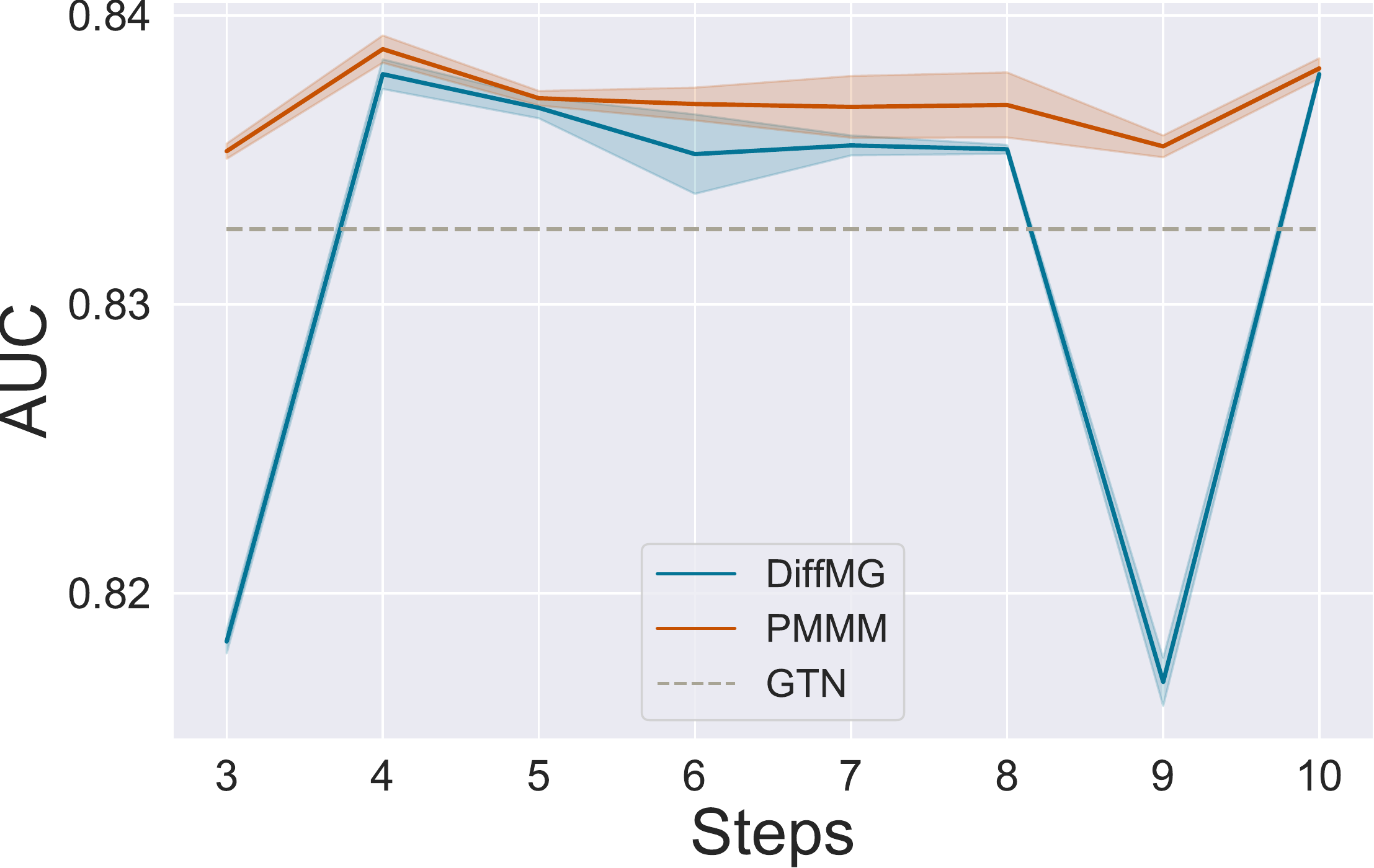}}
\vspace{-4pt}
\caption{Performance on different search steps (best viewed in color). 
}
\vspace{-5pt}
\label{fig:step}
\end{figure*}

\subsection{Visualization}
\label{sec:visual}
Figure~\ref{fig:visual} visualizes the model architectures searched on DBLP for node classification and on Amazon for recommendation. The searched meta-structures by PMMM are more complex than those of DiffMG. However, more message passing steps have little negative impact on efficiency due to parallelization in the training of neural architecture. 
In Figure~\ref{fig:visual} (a), $H^{(2)}$ has two  incoming edge types from $H^{(1)}$. 
We found that both $A_{AP}$ and $A_{PC}$ are critical for message passing between $H^{(1)}$ and $H^{(2)}$. If we search for a meta graph like DiffMG, $A_{AP}$ has to be discarded as multiple paths are not allowed, which will seriously impact the performance. 


\subsection{Stability Study} 
\label{sec:robust}
To evaluate the stability of our method, we compare PMMM with differentiable meta graph search from two perspectives. 

\subsubsection{Random search seeds} 
We evaluate the stability of each algorithm by running the two algorithms on random search seeds from $0$ to $30$, and plot the Macro-F1 scores averaged from $3$ dependent retraining of the searched architecture under different random training seeds. The results are illustrated in Figure~\ref{fig:seed}. The gray dotted line shows the results of hand-designed heterogeneous GNNs, HAN on node classification and GTN on recommendation as the baselines. Although DiffMG shows excellent performance in a few search seeds, the performance dramatically declines in most other seeds. In most cases, its performance is even worse than HAN and GTN. 
In contrast, 
PMMM can overcome the instability issue in DiffMG. Besides, PMMM significantly outperforms DiffMG in most search seeds and consistently surpasses the manual designed networks. Specifically, considering the average results on search seeds from 0 to 30, PMMM has $8.97\%$, $8.33\%$, $1.53\%$, $1.78\%$, $2.67\%$ and $3.11\%$ better performance compared to DiffMG.

\subsubsection{Steps} The results of DiffMG are conducted under the steps (depth) of meta graph $N=4$. We follow their setting for a fair comparison in the main experiments. To explore the performance of the two algorithms on various depths of meta-structures, we run both algorithms in different steps. Figure~\ref{fig:step} illustrates the results. 
The performance of DiffMG is unstable under different depths of meta graphs and often worse than manual networks, which extremely limits its practical application. 

\begin{figure}[!t]
\centering
\subfigure[DBLP]{\includegraphics[width=0.47\linewidth]{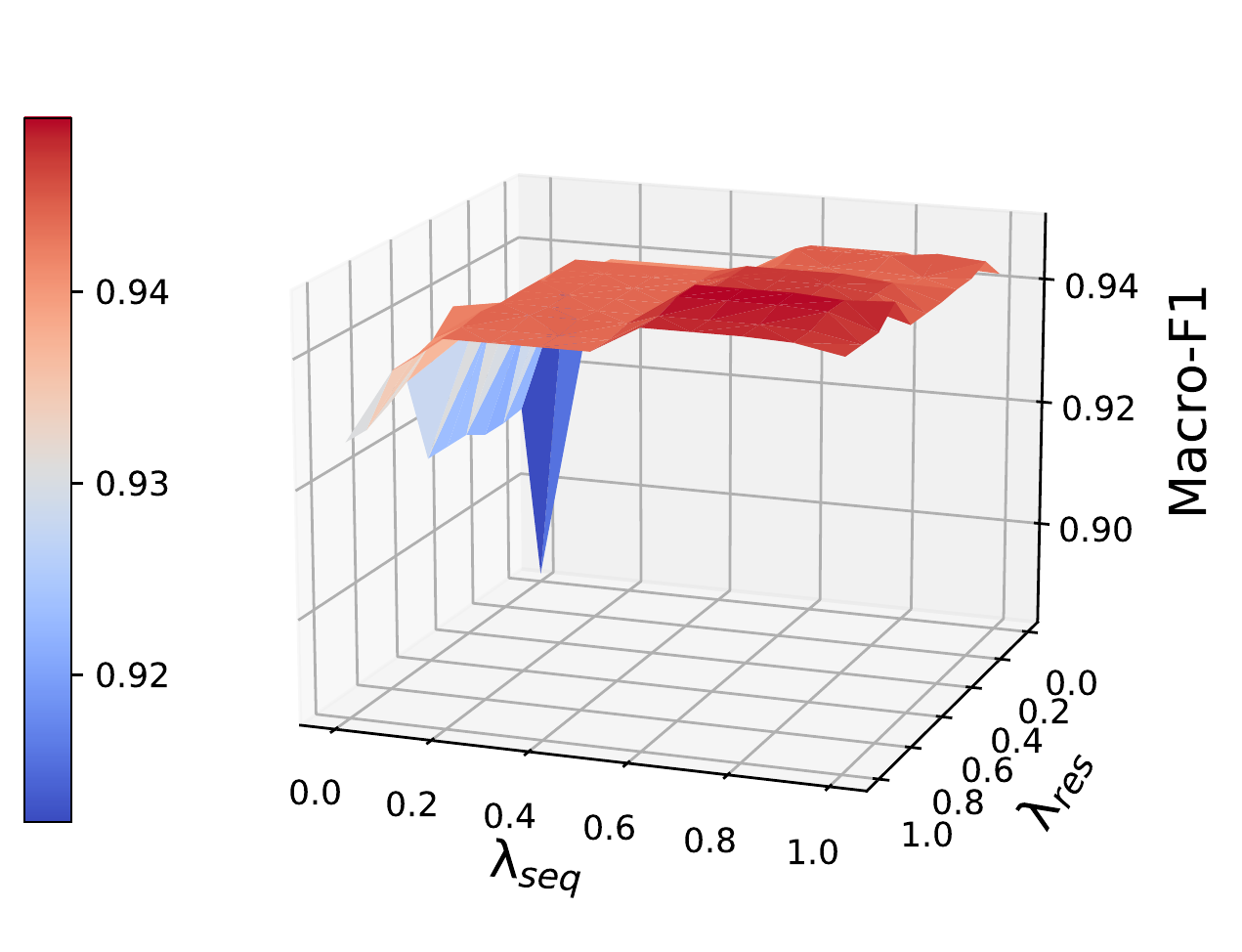}}
\hspace{0.1em}
\subfigure[Amazon]{\includegraphics[width=0.50\linewidth]{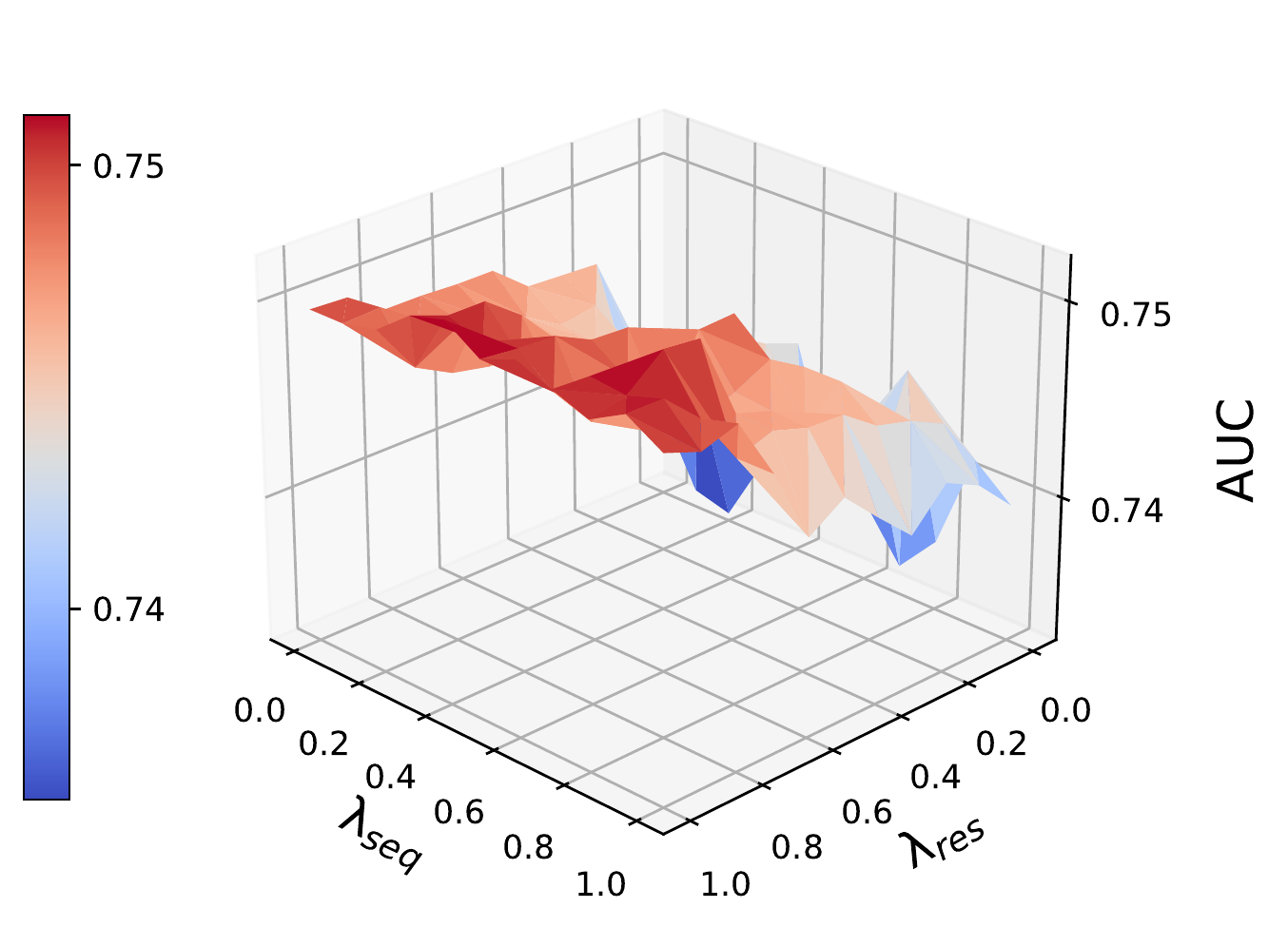}}
\vspace{-5pt}
\caption{Performances of PMMM 
with varying $\lambda$.} 
\vspace{-5pt}
\label{fig:ablation}
\end{figure}
\subsection{Parameter Study} 
We carry out analysis on hyper-parameters $\lambda$ controlling derived meta multigraphs of our model. When $\lambda=1$, only the strongest path is retained in each edge, which is the deriving strategy of most differentiable NAS, including DiffMG. When $\lambda=0$, all paths are retained. We speculate that $\lambda$ should be close to 1 to ensure that all effective paths are retained and weak paths are dropped.
We divide $\lambda$ into two parts, $\lambda_{seq}$ and $\lambda_{res}$, which controls the number of retrained paths in sequential multi-edges and in residual multi-edges, respectively. We vary $\lambda_{seq}$ and $\lambda_{res}$ from $0$ to $1$ 
and plot the results of DBLP for node classification and Amazon for recommendation in Figure~\ref{fig:ablation}. The performance reaches the peak when $\lambda_{seq}$ and $\lambda_{res}$ are around $0.9$, which demonstrates the effectiveness of meta multigraph strategy and verifies our speculation.
Notably, PMMM can derive flexible meta multigraphs by varying $\lambda_{seq}$ and $\lambda_{res}$.


\section{Conclusion}
\label{sec:conclusion}
In this work, we propose a new concept of meta multigraph and present critical contributions to both the first stable search algorithm and flexible deriving strategy for differentiable neural architecture search on HINs. Extensive experiments demonstrate that our method is stable, flexible, and consistently surpassing state-of-the-art heterogeneous GNNs on six datasets of two representative tasks on HINs. In future work, we will try to explore the interpretability of meta multigraphs theoretically.

\section*{Acknowledgments}
This work is supported by National Natural Science Foundation (U22B2017).


\bibliography{aaai23}

\begin{thebibliography}{36}
\providecommand{\natexlab}[1]{#1}

\bibitem[{Anandalingam and Friesz(1992)}]{anandalingam1992hierarchical}
Anandalingam, G.; and Friesz, T.~L. 1992.
\newblock Hierarchical optimization: An introduction.
\newblock \emph{Annals of Operations Research}, 34(1): 1--11.

\bibitem[{Cai, Zhu, and Han(2019)}]{cai2018proxylessnas}
Cai, H.; Zhu, L.; and Han, S. 2019.
\newblock ProxylessNAS: Direct Neural Architecture Search on Target Task and
  Hardware.
\newblock In \emph{7th International Conference on Learning Representations,
  {ICLR} 2019, New Orleans, LA, USA, May 6-9, 2019}.

\bibitem[{Colson, Marcotte, and Savard(2007)}]{colson2007overview}
Colson, B.; Marcotte, P.; and Savard, G. 2007.
\newblock An overview of bilevel optimization.
\newblock \emph{Annals of operations research}, 153(1): 235--256.

\bibitem[{Ding et~al.(2021)Ding, Yao, Zhao, and
  Zhang}]{DBLP:conf/kdd/DingYZZ21}
Ding, Y.; Yao, Q.; Zhao, H.; and Zhang, T. 2021.
\newblock DiffMG: Differentiable Meta Graph Search for Heterogeneous Graph
  Neural Networks.
\newblock In \emph{{KDD} '21: The 27th {ACM} {SIGKDD} Conference on Knowledge
  Discovery and Data Mining, Virtual Event, Singapore, August 14-18}, 279--288.

\bibitem[{Dong, Chawla, and Swami(2017)}]{dong2017metapath2vec}
Dong, Y.; Chawla, N.~V.; and Swami, A. 2017.
\newblock metapath2vec: Scalable Representation Learning for Heterogeneous
  Networks.
\newblock In \emph{{KDD} '17: Proceedings of the 23rd {ACM} {SIGKDD}
  International Conference on Knowledge Discovery and Data Mining, Halifax, NS,
  Canada, August 13 - 17, 2017}, 135--144.

\bibitem[{Fey and Lenssen(2019)}]{Fey/Lenssen/2019}
Fey, M.; and Lenssen, J.~E. 2019.
\newblock Fast Graph Representation Learning with {PyTorch Geometric}.
\newblock In \emph{ICLR Workshop on Representation Learning on Graphs and
  Manifolds}.

\bibitem[{Fu et~al.(2020)Fu, Zhang, Meng, and King}]{fu2020magnn}
Fu, X.; Zhang, J.; Meng, Z.; and King, I. 2020.
\newblock {MAGNN:} Metapath Aggregated Graph Neural Network for Heterogeneous
  Graph Embedding.
\newblock In \emph{{WWW} '20: The Web Conference 2020, Taipei, Taiwan, April
  20-24, 2020}, 2331--2341.

\bibitem[{Gao et~al.(2021)Gao, Zhang, Li, Zhou, Liu, and
  Hu}]{gao2021heterogeneous}
Gao, Y.; Zhang, P.; Li, Z.; Zhou, C.; Liu, Y.; and Hu, Y. 2021.
\newblock Heterogeneous Graph Neural Architecture Search.
\newblock In \emph{2021 IEEE International Conference on Data Mining (ICDM)},
  1066--1071.

\bibitem[{Guo et~al.(2020)Guo, Zhang, Mu, Heng, Liu, Wei, and
  Sun}]{guo2020single}
Guo, Z.; Zhang, X.; Mu, H.; Heng, W.; Liu, Z.; Wei, Y.; and Sun, J. 2020.
\newblock Single path one-shot neural architecture search with uniform
  sampling.
\newblock In \emph{European Conference on Computer Vision, ECCV}, 544--560.

\bibitem[{Han et~al.(2020)Han, Xu, Shi, Shang, Ma, Hui, and
  Li}]{han2020genetic}
Han, Z.; Xu, F.; Shi, J.; Shang, Y.; Ma, H.; Hui, P.; and Li, Y. 2020.
\newblock Genetic Meta-Structure Search for Recommendation on Heterogeneous
  Information Network.
\newblock In \emph{{CIKM} '20: The 29th {ACM} International Conference on
  Information and Knowledge Management, Virtual Event, Ireland, October 19-23,
  2020}, 455--464.

\bibitem[{Hu et~al.(2020)Hu, Dong, Wang, and Sun}]{hu2020heterogeneous}
Hu, Z.; Dong, Y.; Wang, K.; and Sun, Y. 2020.
\newblock Heterogeneous Graph Transformer.
\newblock In \emph{{WWW} '20: The Web Conference 2020, Taipei, Taiwan, April
  20-24, 2020}, 2704--2710.

\bibitem[{Jin et~al.(2020)Jin, Qin, Fang, Du, Zhang, Yu, Zhang, and
  Smola}]{jin2020efficient}
Jin, J.; Qin, J.; Fang, Y.; Du, K.; Zhang, W.; Yu, Y.; Zhang, Z.; and Smola,
  A.~J. 2020.
\newblock An Efficient Neighborhood-based Interaction Model for Recommendation
  on Heterogeneous Graph.
\newblock In \emph{{KDD} '20: The 26th {ACM} {SIGKDD} Conference on Knowledge
  Discovery and Data Mining, Virtual Event, CA, USA, August 23-27, 2020},
  75--84.

\bibitem[{Kipf and Welling(2017)}]{kipf2016semi}
Kipf, T.~N.; and Welling, M. 2017.
\newblock Semi-Supervised Classification with Graph Convolutional Networks.
\newblock In \emph{5th International Conference on Learning Representations,
  {ICLR} 2017, Toulon, France, April 24-26, 2017, Conference Track
  Proceedings}.

\bibitem[{Li et~al.(2021)Li, Jin, Song, Zhu, Shi, and Wang}]{li2021graphmse}
Li, Y.; Jin, Y.; Song, G.; Zhu, Z.; Shi, C.; and Wang, Y. 2021.
\newblock GraphMSE: Efficient Meta-path Selection in Semantically Aligned
  Feature Space for Graph Neural Networks.
\newblock In \emph{Proceedings of the AAAI Conference on Artificial
  Intelligence}, 5, 4206--4214.

\bibitem[{Liu, Simonyan, and Yang(2019)}]{liu2018darts}
Liu, H.; Simonyan, K.; and Yang, Y. 2019.
\newblock {DARTS}: Architecture Search.
\newblock In \emph{ICLR}.

\bibitem[{Qin et~al.(2021)Qin, Wang, Zhang, and Zhu}]{qin2021graph}
Qin, Y.; Wang, X.; Zhang, Z.; and Zhu, W. 2021.
\newblock Graph differentiable architecture search with structure learning.
\newblock \emph{Advances in Neural Information Processing Systems}, 34:
  16860--16872.

\bibitem[{Schlichtkrull et~al.(2018)Schlichtkrull, Kipf, Bloem, van~den Berg,
  Titov, and Welling}]{DBLP:conf/esws/SchlichtkrullKB18}
Schlichtkrull, M.~S.; Kipf, T.~N.; Bloem, P.; van~den Berg, R.; Titov, I.; and
  Welling, M. 2018.
\newblock Modeling Relational Data with Graph Convolutional Networks.
\newblock In \emph{ESWC'2018}.

\bibitem[{Sun et~al.(2011{\natexlab{a}})Sun, Han, Yan, Yu, and
  Wu}]{sun2011pathsim}
Sun, Y.; Han, J.; Yan, X.; Yu, P.~S.; and Wu, T. 2011{\natexlab{a}}.
\newblock Pathsim: Meta path-based top-k similarity search in heterogeneous
  information networks.
\newblock \emph{PVLDB}, 4(11): 992--1003.

\bibitem[{Sun et~al.(2011{\natexlab{b}})Sun, Han, Yan, Yu, and Wu}]{Metapath}
Sun, Y.; Han, J.; Yan, X.; Yu, P.~S.; and Wu, T. 2011{\natexlab{b}}.
\newblock PathSim: Meta Path-Based Top-K Similarity Search in Heterogeneous
  Information Networks.
\newblock \emph{Proceedings of the Vldb Endowment}, 4(11): 992--1003.

\bibitem[{Velickovic et~al.(2018)Velickovic, Cucurull, Casanova, Romero,
  Li{\`{o}}, and Bengio}]{velivckovic2017graph}
Velickovic, P.; Cucurull, G.; Casanova, A.; Romero, A.; Li{\`{o}}, P.; and
  Bengio, Y. 2018.
\newblock Graph Attention Networks.
\newblock In \emph{6th International Conference on Learning Representations,
  {ICLR} 2018, Vancouver, BC, Canada, April 30 - May 3, 2018, Conference Track
  Proceedings}.

\bibitem[{Wang et~al.(2019{\natexlab{a}})Wang, Ji, Shi, Wang, Ye, Cui, and
  Yu}]{DBLP:conf/www/WangJSWYCY19}
Wang, X.; Ji, H.; Shi, C.; Wang, B.; Ye, Y.; Cui, P.; and Yu, P.~S.
  2019{\natexlab{a}}.
\newblock Heterogeneous Graph Attention Network.
\newblock In \emph{The World Wide Web Conference, {WWW} 2019, San Francisco,
  CA, USA, May 13-17, 2019}, 2022--2032.

\bibitem[{Wang et~al.(2019{\natexlab{b}})Wang, Ji, Shi, Wang, Ye, Cui, and
  Yu}]{wang2019heterogeneous}
Wang, X.; Ji, H.; Shi, C.; Wang, B.; Ye, Y.; Cui, P.; and Yu, P.~S.
  2019{\natexlab{b}}.
\newblock Heterogeneous Graph Attention Network.
\newblock In \emph{The World Wide Web Conference, {WWW} 2019, San Francisco,
  CA, USA, May 13-17, 2019}, 2022--2032.

\bibitem[{Wei, Zhao, and He(2022)}]{wei2022designing}
Wei, L.; Zhao, H.; and He, Z. 2022.
\newblock Designing the Topology of Graph Neural Networks: A Novel Feature
  Fusion Perspective.
\newblock In \emph{Proceedings of the ACM Web Conference 2022}, 1381--1391.

\bibitem[{Xie and Yuille(2017)}]{xie2017genetic}
Xie, L.; and Yuille, A.~L. 2017.
\newblock Genetic {CNN}.
\newblock In \emph{{IEEE} International Conference on Computer Vision, {ICCV}
  2017, Venice, Italy, October 22-29, 2017}, 1388--1397.

\bibitem[{Xu et~al.(2020)Xu, Xie, Zhang, Chen, Qi, Tian, and Xiong}]{xu2019pc}
Xu, Y.; Xie, L.; Zhang, X.; Chen, X.; Qi, G.; Tian, Q.; and Xiong, H. 2020.
\newblock {PC-DARTS:} Partial Channel Connections for Memory-Efficient
  Architecture Search.
\newblock In \emph{8th International Conference on Learning Representations,
  {ICLR} 2020, Addis Ababa, Ethiopia, April 26-30, 2020}.

\bibitem[{Xue et~al.(2021)Xue, Wang, Yan, Hu, Yang, and
  Sun}]{xue2021rethinking}
Xue, C.; Wang, X.; Yan, J.; Hu, Y.; Yang, X.; and Sun, K. 2021.
\newblock Rethinking Bi-Level Optimization in Neural Architecture Search: A
  Gibbs Sampling Perspective.
\newblock In \emph{Proceedings of the AAAI Conference on Artificial
  Intelligence}, 12, 10551--10559.

\bibitem[{Yang et~al.(2020)Yang, Xiao, Zhang, Sun, and
  Han}]{yang2020heterogeneous}
Yang, C.; Xiao, Y.; Zhang, Y.; Sun, Y.; and Han, J. 2020.
\newblock Heterogeneous Network Representation Learning: A Unified Framework
  with Survey and Benchmark.
\newblock \emph{IEEE Transactions on Knowledge and Data Engineering}.

\bibitem[{Yao et~al.(2020)Yao, Xu, Tu, and Zhu}]{yao2019differentiable}
Yao, Q.; Xu, J.; Tu, W.-W.; and Zhu, Z. 2020.
\newblock Efficient Neural Architecture Search via Proximal Iterations.
\newblock In \emph{AAAI}.

\bibitem[{Yun et~al.(2019{\natexlab{a}})Yun, Jeong, Kim, Kang, and Kim}]{gt}
Yun, S.; Jeong, M.; Kim, R.; Kang, J.; and Kim, H.~J. 2019{\natexlab{a}}.
\newblock Graph Transformer Networks.
\newblock In \emph{Advances in Neural Information Processing Systems 32: Annual
  Conference on Neural Information Processing Systems 2019, NeurIPS 2019,
  December 8-14, 2019, Vancouver, BC, Canada}, 11960--11970.

\bibitem[{Yun et~al.(2019{\natexlab{b}})Yun, Jeong, Kim, Kang, and
  Kim}]{NIPS2019_9367}
Yun, S.; Jeong, M.; Kim, R.; Kang, J.; and Kim, H.~J. 2019{\natexlab{b}}.
\newblock Graph Transformer Networks.
\newblock In \emph{Advances in Neural Information Processing Systems 32: Annual
  Conference on Neural Information Processing Systems 2019, NeurIPS 2019,
  December 8-14, 2019, Vancouver, BC, Canada}, 11960--11970.

\bibitem[{Zhang et~al.(2019{\natexlab{a}})Zhang, Song, Huang, Swami, and
  Chawla}]{DBLP:conf/kdd/ZhangSHSC19}
Zhang, C.; Song, D.; Huang, C.; Swami, A.; and Chawla, N.~V.
  2019{\natexlab{a}}.
\newblock Heterogeneous Graph Neural Network.
\newblock In \emph{Proceedings of the 25th {ACM} {SIGKDD} International
  Conference on Knowledge Discovery {\&} Data Mining, {KDD} 2019, Anchorage,
  AK, USA, August 4-8, 2019}, 793--803.

\bibitem[{Zhang et~al.(2019{\natexlab{b}})Zhang, Song, Huang, Swami, and
  Chawla}]{zhang2019heterogeneous}
Zhang, C.; Song, D.; Huang, C.; Swami, A.; and Chawla, N.~V.
  2019{\natexlab{b}}.
\newblock Heterogeneous Graph Neural Network.
\newblock In \emph{Proceedings of the 25th {ACM} {SIGKDD} International
  Conference on Knowledge Discovery {\&} Data Mining, {KDD} 2019, Anchorage,
  AK, USA, August 4-8, 2019}, 793--803.

\bibitem[{Zhang et~al.(2019{\natexlab{c}})Zhang, Shi, Zhao, and
  King}]{ijcai2019-0592}
Zhang, J.; Shi, X.; Zhao, S.; and King, I. 2019{\natexlab{c}}.
\newblock {STAR-GCN:} Stacked and Reconstructed Graph Convolutional Networks
  for Recommender Systems.
\newblock In \emph{Proceedings of the Twenty-Eighth International Joint
  Conference on Artificial Intelligence, {IJCAI} 2019, Macao, China, August
  10-16, 2019}, 4264--4270.

\bibitem[{Zhao, Yao, and Tu(2021)}]{zhao2021search}
Zhao, H.; Yao, Q.; and Tu, W. 2021.
\newblock Search to aggregate neighborhood for graph neural network.
\newblock In \emph{ICDE}.

\bibitem[{Zhou et~al.(2019)Zhou, Song, Huang, and Hu}]{zhou2019auto}
Zhou, K.; Song, Q.; Huang, X.; and Hu, X. 2019.
\newblock {Auto-GNN}: Neural architecture search of graph neural networks.
\newblock \emph{arXiv preprint arXiv:1909.03184}.

\bibitem[{Zoph and Le(2017)}]{zoph2016neural}
Zoph, B.; and Le, Q.~V. 2017.
\newblock Neural Architecture Search with Reinforcement Learning.
\newblock In \emph{5th International Conference on Learning Representations,
  {ICLR} 2017, Toulon, France, April 24-26, 2017, Conference Track
  Proceedings}.

\end{thebibliography}

\clearpage
\appendix
\section{Appendix}
\label{sec:appendix}
This supplementary material provides additional details and results not included in the main text due to space limitations. We begin by displaying the specifics of all Datasets involved in our experiments of node classification and recommendation. Then, we show the details and parameter settings of the compared baselines. Next, we compare the proposed PMMM with HGNAS~\cite{gao2021heterogeneous} using its data division. Finally, we compare several baselines to evaluate the evaluation efficiency of our method. 

\subsection{Statistics of Datasets}
We evaluate our method on two popular tasks~\cite{yang2020heterogeneous}: node classification and recommendation. In the node classification task, we use three widely-used real-world datasets with huge heterogeneous information: DBLP, ACM, and IMDB.
DBLP contains three types of nodes (papers (P), authors (A), conferences (C)), and four types of edges (PA, AP, PC, CP), and authors are labeled by their research areas. ACM contains three types of nodes (papers (P), authors (A), subject (S)), and four types of edges (PA, AP, PS, SP), and papers are labeled by categories. IMDB contains three types of nodes (movies (M), actors (A), directors (D)), and four types of edges (MA, AM, MD, DM), and nodes are labeled by their movie categories. Nodes in these datasets have bag-of-words representations of graphs as input features. 
We follow the splits in GTN~\cite{NIPS2019_9367} and  DiffMG~\cite{DBLP:conf/kdd/DingYZZ21}. The statistics of the three datasets are shown in Table~\ref{tab:s_nc}.

In the recommendation task, we adopt three widely used heterogeneous recommendation datasets\footnote{\url{https://github.com/librahu/HIN-Datasets-for-Recommendation-and-Network-Embedding}}
: Amazon, Yelp, and Douban Movie. 
Amazon is one of the biggest e-commerce platforms with global operations. 
Yelp is a platform for users to rate and review local businesses. 
Douban Movie is a well-known social media community in China.   
Following the conventions~\cite{DBLP:conf/kdd/DingYZZ21}, we convert ratings into binary class labels by randomly selecting $50\%$ of ratings greater than three as positive pairs, and all the ratings lower than four as negative pairs. The positive pairs were then randomly divided into a training set, a validation set, and a test set in a 3:1:1 ratio. To fairly compare, we randomly split the negative pairs to balance the number of two pairs in each set, and we use one-hot IDs as input features. To avoid the label leakage issue~\cite{ijcai2019-0592}, both positive and negative pairs are disconnected in the original network. The statistics of the three datasets are illustrated in Table~\ref{tab:s_re}.
 
\begin{table}[tp]
    \small
	\centering
	\caption{Statistics of HINs for node classification.}
    \label{tab:s_nc}
    \begin{threeparttable}[b]
	\small
	\resizebox{0.4\textwidth}{!}{
	\begin{tabular}{>{\rowmac}l>{\rowmac}c>{\rowmac}c>{\rowmac}c}
		\toprule
		\setrow{\bfseries}& DBLP   & ACM    & IMDB\clearrow    \\ 
		\midrule
		\#~Nodes        & 18405  & 8994   & 12624   \\
		\#~Edges        & 67946  & 25922  & 37288   \\
		\#~Classes      & 4      & 3      & 3       \\
		\#~Edge types \qquad \qquad   & \ 4  \qquad     & \ 4  \qquad     & 4  \\
		\#~Training     & 800    & 600    & 300     \\
		\#~Validation   & 400    & 300    & 300     \\
		\#~Testing      & 2857   & 2125   & 2339    \\ 
		\bottomrule
	\end{tabular}}
	\end{threeparttable}
\end{table}


\begin{table}[!t]
	\small
	\centering
	\caption{Statistics of HINs for recommendation.}
    \label{tab:s_re}
	\begin{threeparttable}[b]
	\small
	\resizebox{0.48 \textwidth}{!}{
	\begin{tabular}{>{\rowmac}c>{\rowmac}c>{\rowmac}c>{\rowmac}c>{\rowmac}c}
		\toprule
		\setrow{\bfseries}Dataset                   &  Relations (A-B)                & \#~A           & \#~B             & \#~A-B\clearrow           \\ \midrule
		\multirow{5}{*}{Yelp}     &  \textbf{User-Business (U-B)}   & \textbf{16239} & \textbf{14284}   & \textbf{198397}  \\
		~                         &  User-User (U-U)                & 16239          & 16239            & 158590           \\
		~                         &  User-Compliment (U-Co)         & 16239          & 11               & 76875            \\
		~                         &  Business-City (B-C)            & 14284          & 47               & 14267            \\
		~                         &  Business-Category (B-Ca)       & 14284          & 511              & 40009            \\ \midrule
		\multirow{6}{*}{\shortstack{Douban\\movie}} & \textbf{User-Movie (U-M)}  & \textbf{13367} & \textbf{12677} & \textbf{1068278} \\ 
		~   & User-Group (U-G) & 13367 & 2753  & 570047 \\
		~   & User-User  (U-U) & 13367 & 13367 & 4085  \\
		~   & Movie-Actor (M-A) & 12677 & 6311 & 33587 \\
		~   & Movie-Director (M-D) & 12677 & 2449 & 11276 \\
		~   & Movie-Type  (M-T) & 12677 & 38 & 27668 \\ \midrule
		\multirow{4}{*}{Amazon} & \textbf{User-Item (U-I)} & \textbf{6170} & \textbf{2753} & \textbf{195791} \\
		~ & Item-View (I-V) & 2753 & 3857 & 5694 \\
		~ & Item-Category (I-C) & 2753 & 22 & 5508 \\
		~ & Item-Brand (I-B) & 2753 & 334 & 2753 \\ \bottomrule
	\end{tabular}
	}
	\end{threeparttable}
\end{table}

\subsection{Baselines and Hyper-parameters}
\begin{table}[t]
\small
	\caption{Classification of heterogeneous GNNs based on the utilized semantic information.}
	\label{tab:hingnn}
	\begin{threeparttable}[b]
    \footnotesize
    \resizebox{0.45\textwidth}{!}{
	\begin{tabular}{ c |C{130px}}
	\toprule
	Semantics & Method        \\
	\midrule
	Meta path & HAN~\cite{wang2019heterogeneous},\ \ \ 
	GTN~\cite{NIPS2019_9367}, \ \ \ 
	MAGNN~\cite{fu2020magnn}, \ \ \
	HGT~\cite{hu2020heterogeneous}\\
	\midrule
	Meta graph & GEMS~\cite{han2020genetic},\ \ \ 
	DiffMG~\cite{DBLP:conf/kdd/DingYZZ21} \\
    \midrule
	Meta Multigraph & \model\\
	\bottomrule
	\end{tabular}}
	\end{threeparttable}
\end{table}

We compare our method with eleven methods, including 
one random walk based network embedding method,  
two homogeneous GNNs, 
five heterogeneous GNNs , 
three AutoML methods. Table~\ref{tab:hingnn} summarizes all heterogeneous GNN baselines and our own model. In our experiments, all GNN methods are trained using a full batch. We set the training epoch to 100 in the node classification task with early stopping and 200 in the recommendation task.  
For a fair comparison, the hyper-parameters of baselines simply follow the setting in DiffMG~\cite{DBLP:conf/kdd/DingYZZ21}.
The list of baselines and important hyper-parameters settings are shown as follows. Other hyper-parameters follow the default settings.
\begin{itemize} 
\item \textbf{metapath2vec}~\cite{dong2017metapath2vec}\footnote{\url{https://ericdongyx.github.io/metapath2vec/m2v.html}} is a traditional heterogeneous network embedding model. It generates random walks guided by one single meta-path and utilizes the skip-gram model to perform heterogeneous graph embedding. In our experiments, we follow its default hyper-parameter settings.
\item \textbf{GCN}~\cite{kipf2016semi}\footnote{We use implementations from PyTorch Geometric (PyG)~\cite{Fey/Lenssen/2019} for GCN and GAT.} is a homogeneous GNN. It creates a homophily degree matrix using label propagation to explore the extent to which a pair of nodes belong to the same class. In our experiments, we use $4$ layers for GCN, and the hidden dimension is set to $64$ for a fair comparison. 
\item \textbf{GAT}~\cite{velivckovic2017graph} is a homogeneous GNN. This model learns the influence weights of each neighbor via a self-attention mechanism, ignoring the heterogeneity of nodes. In our experiments, we use $4$ layers for GAT, and the hidden dimension is set to $64$. The number of attention heads is $8$.
\item \textbf{HAN}~\cite{wang2019heterogeneous}\footnote{We implement with PyG following the authors' TensorFlow code at \url{https://github.com/Jhy1993/HAN}.} is a heterogeneous GNN. It converts a heterogeneous graph into multiple homogeneous graphs constructed by meta-paths and leverages the attention mechanism on these graphs. In our experiments, the hidden dimension is set to $64$, and the number of attention heads is $8$. The dimension of the semantic-level attention vector is $128$.
\item \textbf{MAGNN}~\cite{fu2020magnn}\footnote{\url{https://github.com/cynricfu/MAGNN}} is a heterogeneous GNN. This model applies intra-meta-path aggregation for each meta-path and inter-meta-path aggregation. In our experiments, the hidden dimension is set to $64$, and the number of attention heads is $8$. The dimension of the semantic-level attention vector is $128$.
\item \textbf{GTN}~\cite{NIPS2019_9367}\footnote{\url{https://github.com/seongjunyun/Graph_Transformer_Networks}} is a heterogeneous GNN. It generates meta-paths through stacking multiple graph transformer layers and subsequent graph convolution. In our experiments, the number of attention heads is $8$.
\item \textbf{HGT}~\cite{hu2020heterogeneous}\footnote{\url{https://github.com/acbull/pyHGT}} is a heterogeneous GNN. It uses the meta relations of heterogeneous graphs to parameterize weight matrices and learns the implicit meta paths. For HGT, we use $4$ layers, and the hidden dimension is set to $64$. The number of attention heads is $8$.
\item \textbf{GraphMSE}~\cite{li2021graphmse}\footnote{\url{https://github.com/pkuliyi2015/GraphMSE}} is a heterogeneous GNN. It combines GNNs and automatic meta-path selection. It can automatically sample meta-path instances and employs GNN with embeddings of these instances. Since DBLP, ACM, and IMDB datasets are also used in their experiments, we directly apply the corresponding settings in their papers. 
\item \textbf{HGNAS}~\cite{gao2021heterogeneous} is an AutoML method searching for heterogeneous GNN. It searches for message encoding and aggregation functions instead of meta graphs by using reinforcement learning. HGNAS employs a different data division and does not provide the source code, so we do a separate comparison with HGNAS using its data division.
\item \textbf{GEMS}~\cite{han2020genetic}\footnote{\url{https://github.com/0oshowero0/GEMS}} is an AutoML method searching for heterogeneous GNN. It utilizes the evolutionary algorithm to search for meta graphs for recommendation on HINs. For GEMS, we run the evolutionary algorithm for 100 generations with a population size of 20.
\item \textbf{DiffMG}~\cite{DBLP:conf/kdd/DingYZZ21}\footnote{\url{https://github.com/AutoML-Research/DiffMG}} is an AutoML method searching for heterogeneous GNN. It employs an efficient differentiable algorithm to search for meta graphs on HINs. DiffMG is the most related work to our method, so we directly follow the setting in DiffMG.
\end{itemize}

\subsection{Comparison with HGNAS}
Besides GEMS~\cite{han2020genetic} and DiffMG~\cite{DBLP:conf/kdd/DingYZZ21}, HGNAS~\cite{gao2021heterogeneous} also employs NAS in HINs.
It searches for message encoding and aggregation functions instead of meta graphs by using reinforcement learning. However, it employs a data division different from ours and doesn't provide the source code. So we compare PMMM with it using its data division in this subsection.
\begin{table}[!t]
	\centering
	\caption{ Macro-F1 ($\%$) of HGNAS and PMMM on node classification.}
	\label{tab:comparison}
	\begin{threeparttable}[b]
    \footnotesize
    \resizebox{0.38\textwidth}{!}{
	\begin{tabular}{ccccc}
		\toprule
		\multirow{2}{*}{}& \multicolumn{3}{c}{\textbf{HGNAS}}& \multirow{2}{*}{\textbf{PMMM}} \\
		& $20\%$& $40\%$& $60\%$& \\
		\midrule
        DBLP     &93.93   &93.94    &94.48 & \textbf{94.61}   \\  
	    ACM      & 91.56  & 91.87  & 92.03 & \textbf{92.58}\\ 
        \bottomrule
	\end{tabular}}
	\end{threeparttable}
\end{table}
Following HGNAS, we show the performance of PMMM on DBLP and ACM. The author nodes of DBLP are divided into training, validation, and testing sets of 400 (9.86\%), 400 (9.86\%), and 3257 (80.28\%) nodes, respectively. The movie nodes are divided into training, validation, and testing sets of 400 (9.35\%), 400 (9.35\%), and 3478 (81.30\%) nodes, respectively. HGNAS feeds the embeddings of labeled nodes (papers in ACM and authors in DBLP) generated by each learning model to an SVM classifier with varying training proportions (\ie, $20\%$) to enhance the performance. In contrast, PMMM doesn't employ an SVM classifier.
As illustrated in Table~\ref{tab:comparison}, PMMM performs consistently better than all results of HGNAS even if it uses an additional SVM classifier.

\subsection{Evaluation Efficiency}
\begin{figure}[!t]
\centering
\subfigure[Yelp]{\includegraphics[width=0.48\linewidth]{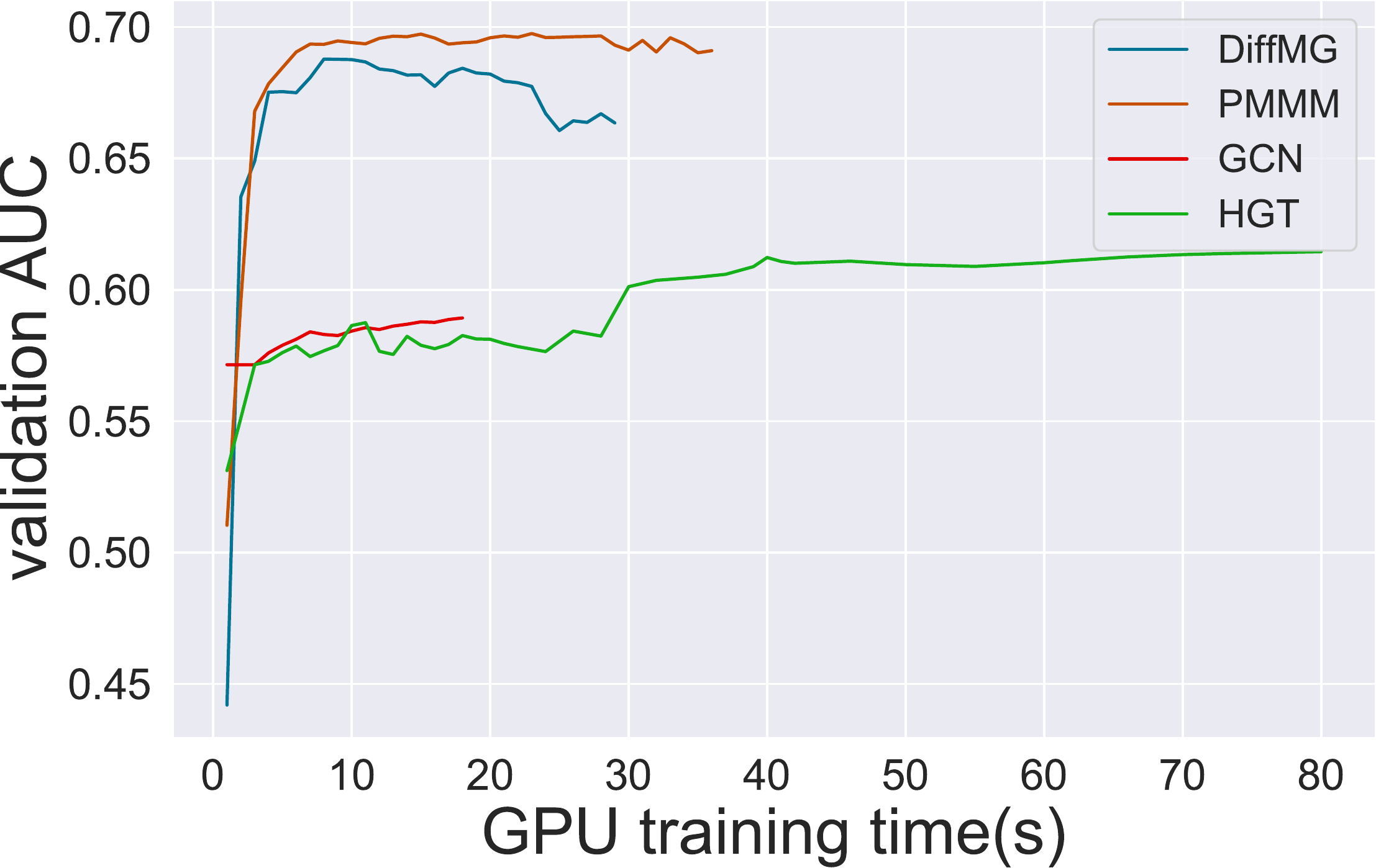}}
\hspace{0.1em}
\subfigure[Douban]{\includegraphics[width=0.48\linewidth]{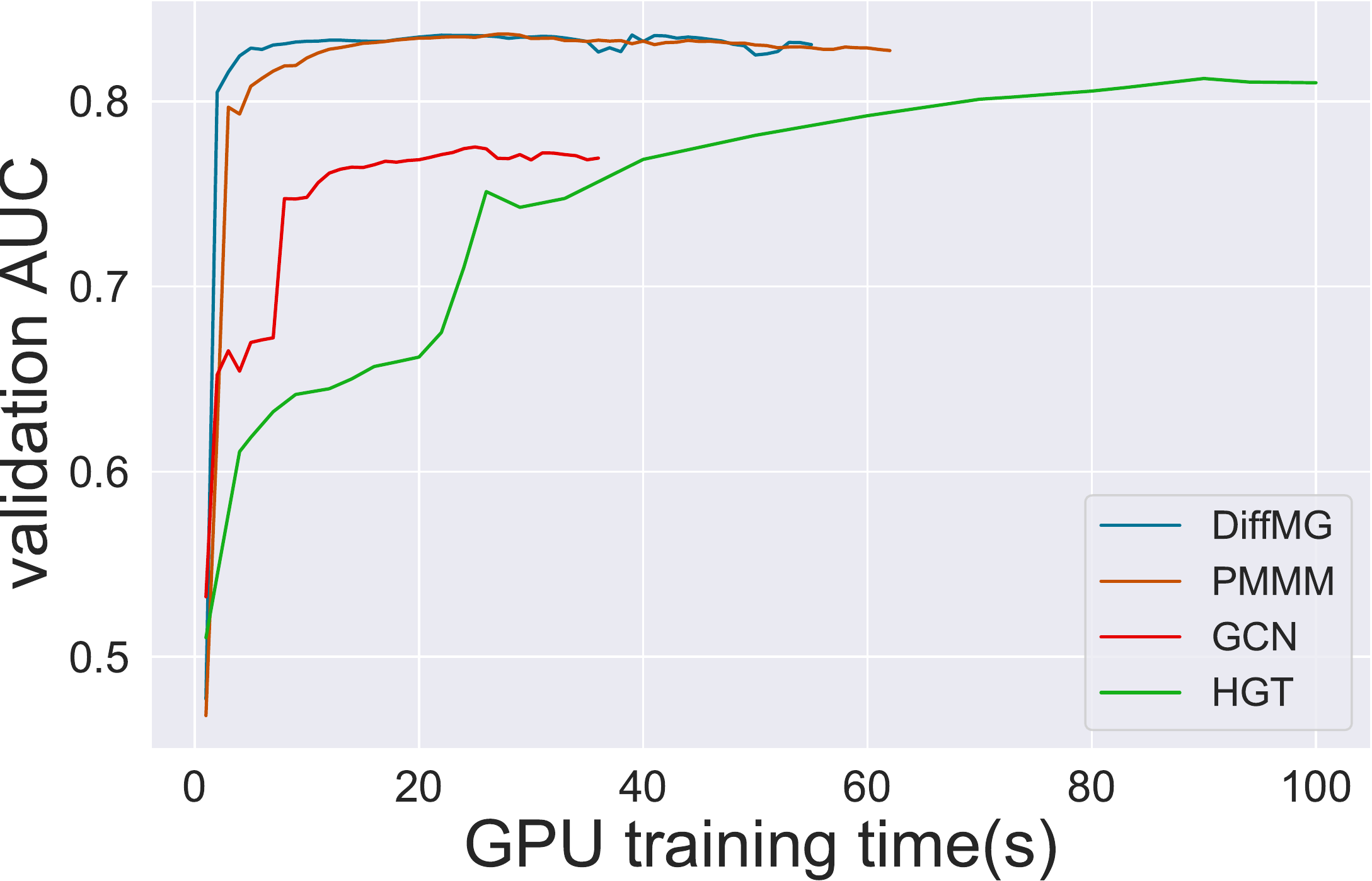}}
\caption{Efficiency of PMMM at the evaluation stage. 
}
\label{fig:efficiency_eval}
\end{figure}
We evaluate the efficiency of our method at the evaluation stage on two large heterogeneous datasets: Yelp and Douban. GCN is the simplest architecture in GNNs. We compare PMMM with it to demonstrate the trade-off between performance and efficiency. HGT outperforms the other heterogeneous GNNs on recommendation task. DiffMG is the most related NAS approach to ours. So we compare against GCN~\cite{kipf2016semi}, HGT~\cite{hu2020heterogeneous} and DiffMG.
Another strong baseline GTN~\cite{NIPS2019_9367} takes a much longer time to train on CPU (nearly 20 minutes per epoch on Yelp) and cannot fit into a single GPU on these datasets, so it is not compared here.

We plot the validation AUC versus GPU training time (measured in seconds) of 200 epochs in Figure~\ref{fig:efficiency_eval}. Similar to DiffMG, PMMM achieves its peak validation performance much faster than HGT. Compared with GCN, PMMM gains large performance improvement while taking only a minor amount of extra time to complete training.
Compared with DiffMG, our method takes little additional time, which is negligible given the improvement in test performance and stability shown in the main text.


\end{document}